%% file: main.tex
\documentclass[10pt,twocolumn,letterpaper]{article}

\usepackage{cvpr}
\usepackage{times}
\usepackage{epsfig}
\usepackage{graphicx}
\usepackage{amsmath}
\usepackage{amssymb}
\usepackage{float}
\usepackage{tabularx}

\usepackage{pgfplots}
  \pgfplotsset{compat=newest}
\usetikzlibrary{plotmarks}
\usepackage{grffile}
\usepackage{amsmath}

\usepackage{overpic}
\usepackage{pgf,tikz}
\usepackage{tkz-euclide}
\usepackage{tkz-graph}
\usetikzlibrary{calc}
\usetkzobj{all}

\usepackage{pgfplots}
\newlength\figureheight 
\newlength\figurewidth

\usepackage{booktabs}
\usepackage{multirow}

\usepackage{hyperref}

\cvprfinalcopy 

\def\cvprPaperID{2123} 

\ifcvprfinal\fi
\begin{document}

\title{Geometric deep learning on graphs and manifolds using mixture model CNNs}

\author{
Federico Monti$^{1} \thanks{Equal contribution}$ 
\hspace{1cm}
Davide Boscaini$^{1 \ast}$
\hspace{1cm}
Jonathan Masci$^{1,4}$ \vspace{1mm}
\\ 
\hspace{5mm}
Emanuele Rodol{\`a}$^1$
\hspace{1.2cm}
Jan Svoboda$^1$
\hspace{0.9cm}
Michael M. Bronstein$^{1,2,3}$ \vspace{1mm}
\\ 
{
$^1$USI Lugano \hspace{5mm}
$^2$Tel Aviv University \hspace{5mm}
$^3$Intel Perceptual Computing \hspace{5mm}
$^4$Nnaisense
}
}


\makeatletter
\renewcommand{\paragraph}{%
  \@startsection{paragraph}{4}%
  {\z@}{2.0ex \@plus 1ex \@minus .2ex}{-1em}%
  {\normalfont\normalsize\bfseries}%
}
\makeatother

\maketitle

\begin{abstract}
Deep learning has achieved a remarkable  performance breakthrough in several fields, most notably in speech recognition, natural language processing, and computer vision. In particular, convolutional neural network (CNN) architectures currently produce state-of-the-art performance on a variety of image analysis tasks such as object detection and recognition. 
Most of deep learning research has so far focused on dealing with 1D, 2D, or 3D Euclidean-structured data such as acoustic signals, images, or videos. 
Recently, there has been an increasing interest in {\em geometric deep learning}, attempting to generalize deep learning methods to non-Euclidean structured data such as graphs and manifolds, with a variety of applications from the domains of network analysis, computational social science, or computer graphics. 
In this paper, we propose a unified framework allowing to generalize CNN architectures to non-Euclidean domains  (graphs and manifolds) and learn local, stationary, and compositional task-specific features. 
We show that various non-Euclidean CNN methods previously proposed in the literature can be considered as particular instances of our framework. 
We test the proposed method on standard tasks from the realms of image-, graph- and 3D shape analysis and show that it consistently outperforms previous approaches.

\end{abstract}

\vspace{-2mm}
\input{intro}
%
%
\input{dl_graphs}

\input{dl_manifolds}

\input{our_method}

\section{Results}

\input{results_images}

\input{results_graphs}

\input{results_shapes}

\input{conclusions}

\section*{Acknowledgments}
This research was supported in part by the ERC Starting Grant No. 307047 (COMET), a Google Faculty Research Award, and Nvidia equipment grant.

{\small
\bibliographystyle{ieee}
\bibliography{refs}
}

\end{document}

%% file: intro.tex
\section{Introduction}

In recent years, increasingly more fields have to deal with geometric non-Euclidean structured data such as manifolds or graphs. 
Social networks are perhaps the most prominent example of such data; additional examples include transportation networks, sensor networks, functional networks representing anatomical and functional structure of the brain, and regulatory networks modeling gene expressions. 
In computer graphics, 3D objects are traditionally modeled as Riemannian manifolds. 
The success of deep learning methods in many fields has recently provoked a keen interest in {\em geometric deep learning} \cite{review_new} attempting to generalize such methods to non-Euclidean structure data.

\subsection{Related works}
\label{sec:rel-works}

\paragraph*{Image processing.} 
Classical deep learning algorithms build on top of traditional signal processing that has been developed primarily for linear shift-invariant systems, naturally arising when dealing with signals on Euclidean spaces. In this framework, basic filtering operations are represented as convolutions. A significant paradigm shift in image processing came with the pioneering work of Perona and Malik \cite{perona1990scale}, suggesting the use of non-shift-invariant image filtering preserving the edge structures. This work was the precursor of a whole new class of PDE-based methods for image processing. Sochen \etal \cite{sochen1998general} brought geometric models into image processing, considering images as manifolds and employing tools from differential geometry for their processing and analysis. More recent graph-based image processing methods relying on spectral graph theory \cite{shi2000normalized,sanfeliu2002graph,zhang2008graph,lezoray2012image} can be traced back to these works. 

\paragraph*{Manifold learning.} 
A similar trend of employing geometric models can also be observed in the machine learning community in the past decade. Modelling data as low-dimensional manifolds is the core of manifold learning techniques such as Laplacian eigenmaps \cite{belkin2003laplacian} for non-linear dimensionality reduction, spectral clustering \cite{ng2002spectral} or spectral hashing \cite{weiss2009spectral}.  
\paragraph*{Signal processing on graphs.}
More recent works tried to generalize signal processing methods to graph-based data  \cite{shuman2013emerging}. Spectral analysis techniques were extended to graphs considering the orthogonal eigenfunctions of the Laplacian operator as a generalization of the Fourier basis. Constructions such as wavelets \cite{coifman2006diffusionw,gavish2010multiscale,hammond2011wavelets,rustamov2013wavelets,sharon2015class} or algorithms such as dictionary learning \cite{zhang2012learning}, Lasso \cite{bresson2015enhanced}, PCA \cite{shahid2015robust,shahid2016fast}, or matrix completion \cite{kalofolias2014matrix} originally developed for the Euclidean domain, were also applied to graph-structured data.

\paragraph*{Deep learning on graphs.}
%
%
The earliest attempts to generalize neural networks to graphs we are aware of are due to Scarselli \etal \cite{gori2005new,GNN}. This work remained practically unnoticed and has been rediscovered only recently \cite{GGSNN,comnets}. 
The interest in non-Euclidean deep learning has recently surged in the computer vision and machine learning communities after the seminal work of Bruna \etal \cite{bruna2013spectral,henaff2015deep}, in which the authors formulated CNN-like \cite{lecun1998gradient} deep neural architectures on graphs in the spectral domain, employing the analogy between the classical Fourier transforms and projections onto the eigenbasis of the graph Laplacian operator \cite{shuman2013emerging}. 
In a follow-up work, Defferrard \etal \cite{defferrard2016convolutional} proposed an efficient filtering scheme that does not require explicit computation of the Laplacian eigenvectors by using recurrent Chebyshev polynomials. 
Kipf and Welling \cite{welling2016} further simplified this approach using simple filters operating on 1-hop neighborhoods of the graph.
Similar methods were proposed in 
\cite{atwood2016search} and \cite{duvenaud2015convolutional}.
%
%
Finally, in the network analysis community, several works constructed graph embeddings \cite{perozzi2014deepwalk,tang2015line,cao2015grarep,grovernode2vec,yang2016revisiting} methods inspired by the Word2Vec technique \cite{mikolov2013distributed}.

A key criticism of spectral approaches such as \cite{bruna2013spectral,henaff2015deep,defferrard2016convolutional} is the fact that the spectral definition of convolution is dependent on the Fourier basis (Laplacian eigenbasis), which, in turn is domain-dependent. It implies that a spectral CNN model learned on one graph cannot be trivially transferred to another graph with a different Fourier basis, as it would be expressed in a `different language'.

\paragraph*{Deep learning on manifolds.}
In the computer graphics community, we can notice a parallel effort of generalizing deep learning architectures to 3D shapes modeled as manifolds (surfaces). 
Masci \etal \cite{masci2015geodesic} proposed the first intrinsic version of convolutional neural networks on manifolds applying filters to local patches represented in geodesic polar coordinates \cite{isc}. Boscaini \etal \cite{boscaini2016learning} used anisotropic heat kernels as an alternative way of extracting intrinsic patches on manifolds. 
In \cite{boscaini2015learning}, the same authors proposed a CNN-type architecture in the spatio-frequency domain using the windowed Fourier transform formalism \cite{shuman2016vertex}. 
Sinha \etal \cite{sinha2016deep} used geometry images representation to obtain Euclidean parametrization of 3D shapes on which standard CNNs can be applied.

The key advantage of spatial techniques is that they generalize across different domains, which is a crucial property in computer graphics applications (where a CNN model can be trained on one shape and applied to another one). 
However, while spatial constructions such as anisotropic heat kernels have a clear geometric interpretation on manifolds, their interpretation on general graphs is somewhat elusive.


\subsection{Main contribution}

In this paper, we present {\em mixture model networks} (MoNet), a general framework allowing to design convolutional deep architectures on non-Euclidean domains such as graphs and manifolds. 
Our approach follows the general philosophy of spatial-domain methods such as \cite{masci2015geodesic,boscaini2016learning,atwood2016search}, formulating convolution-like operations as template matching with local intrinsic `patches' on graphs or manifolds. 
The key novelty is in the way in which the patch is extracted: while previous approaches used fixed patches, e.g. in geodesic or diffusion coordinates, we use a parametric construction. 
In particular, we show that patch operators can be constructed as a function of local graph or manifolds pseudo-coordinates, and study a family of functions represented as a mixture of Gaussian kernels.  
Such a construction allows to formulate previously proposed Geodesic CNN (GCNN) \cite{masci2015geodesic} and Anisotropic CNN (ACNN) \cite{boscaini2016learning} on manifolds or GCN \cite{welling2016} and DCNN \cite{atwood2016search} on graphs as particular instances of our approach.

Among applications on which we exemplify our approach are classical problems from the realms of image-, graph- and 3D- shape analysis. 
In the first class of problems, the task is to classify images, treated as adjacency graphs of superpixels. 
In the second class of problems, we perform vertex-wise classification on a graph representing a citation network of scientific papers. 
Finally, we consider the problem of finding dense intrinsic correspondence between 3D shapes, treated as manifolds. 
In all the above problems, we show that our approach consistently outperforms previously proposed non-Euclidean deep learning methods. 
%




%% file: dl_graphs.tex
\section{Deep learning on graphs}
\label{sec:graph_cnn}

Let $\mathcal{G} = (\{1,\hdots, n\}, \mathcal{E}, \mathbf{W})$ be an undirected weighted graph, represented by the {\em adjacency matrix} $\mathbf{W} = (w_{ij})$, where $w_{ij} = w_{ji}$, $w_{ij} = 0$ if $(i,j) \notin \mathcal{E}$ and $w_{ij} > 0$ if $(i,j) \in \mathcal{E}$. 
The (unnormalized) {\em graph Laplacian} is an $n\times n$ symmetric positive-semidefinite matrix $\boldsymbol{\Delta} = \mathbf{D} - \mathbf{W}$, where $\mathbf{D} = \mathrm{diag}\left(\sum_{j\neq i} w_{ij} \right)$ is the {\em degree matrix}.

The Laplacian has an eigendecomposition 
$
\boldsymbol{\Delta} = \boldsymbol{\Phi} \boldsymbol{\Lambda} \boldsymbol{\Phi}^\top
$, 
where $\boldsymbol{\Phi} = (\boldsymbol{\phi}_1, \hdots \boldsymbol{\phi}_n)$ are the orthonormal eigenvectors and $\boldsymbol{\Lambda} = \mathrm{diag}(\lambda_1, \hdots, \lambda_n)$ is the diagonal matrix of corresponding eigenvalues. The eigenvectors play the role of Fourier atoms in classical harmonic analysis and the eigenvalues can be interpreted as frequencies. 
Given a signal $\mathbf{f} = (f_1, \hdots, f_n)^\top$ on the vertices of graph $\mathcal{G}$, its {\em graph Fourier transform} is given by $\hat{\mathbf{f}} = \boldsymbol{\Phi}^\top\mathbf{f}$. 
Given two signals $\mathbf{f}, \mathbf{g}$ on the graph, their {\em spectral convolution} can be defined as the element-wise product of the  Fourier transforms,
\begin{equation} 
\label{spectral_conv}
\mathbf{f} \star \mathbf{g} = \boldsymbol{\Phi}(\boldsymbol{\Phi}^\top\mathbf{f}) \circ (\boldsymbol{\Phi}^\top\mathbf{g}) = \boldsymbol{\Phi}\, \mathrm{diag}(\hat{g}_1, \hdots, \hat{g}_n)\hat{\mathbf{f}},
\end{equation}
which corresponds to the property referred to as the {\em Convolution Theorem} in the Euclidean case. 


\paragraph*{Spectral CNN.} 
Bruna \etal \cite{bruna2013spectral} used the spectral definition of convolution~(\ref{spectral_conv}) to generalize CNNs on graphs, with a spectral convolutional layer of the form 
\begin{equation} 
\label{spectral_construction_eq}
\mathbf{f}^{\mathrm{out}}_l =   \xi \left(  \sum_{l'=1}^{p} \boldsymbol{\Phi}_k \hat{\mathbf{G}}_{l,l'} \boldsymbol{\Phi}_k^\top \mathbf{f}^{\mathrm{in}}_{l'} \right).
\end{equation}
Here the $n\times p$ and $n\times q$ matrices $\mathbf{F}^{\mathrm{in}} = (\mathbf{f}^{\mathrm{in}}_1, \hdots, \mathbf{f}^{\mathrm{in}}_p)$  and $\mathbf{F}^{\mathrm{out}} = (\mathbf{f}^{\mathrm{out}}_1, \hdots, \mathbf{f}^{\mathrm{out}}_q)$ represent respectively the $p$- and $q$-dimensional input and output signals on the vertices of the graph, 
$\boldsymbol{\Phi} = (\boldsymbol{\phi}_1, \hdots, \boldsymbol{\phi}_k)$ is an $n\times k$ matrix of the first eigenvectors, 
$\hat{\mathbf{G}}_{l,l'} = \mathrm{diag}(\hat{g}_{l,l',1}, \hdots, \hat{g}_{l,l',k})$ is a $k\times k$ diagonal matrix of spectral multipliers representing a learnable filter in the frequency domain, and $\xi$ is a nonlinearity (e.g. ReLU) applied on the vertex-wise function values. 
%
%
%
%
The analogy of pooling in this framework is a graph coarsening procedure, which, given a graph with $n$ vertices, produces a graph with $n'<n$ vertices and transfers signals from the vertices of the fine graph to those of the coarse one.

While conceptually important, this framework has several major drawbacks. 
First, the spectral filter coefficients are {\em basis dependent}, and consequently, a spectral CNN model learned on one graph cannot be applied to another graph. 
Second, the computation of the forward and inverse graph Fourier transform incurs expensive $\mathcal{O}(n^2)$ multiplication by the matrices $\boldsymbol{\Phi}, \boldsymbol{\Phi}^\top$, as there is no FFT-like algorithms on general graphs. 
Third, there is no guarantee that the filters represented in the spectral domain are localized in the spatial domain; assuming $k = O(n)$ eigenvectors of the Laplacian are used, a spectral convolutional layer requires 
$p q k = O(n)$ parameters to train.

\paragraph*{Smooth Spectral CNN.} 
In a follow-up work, Henaff \etal \cite{henaff2015deep} argued that {\em smooth} spectral filter coefficients result in spatially-localized filters and used parametric filters of the form
\begin{equation}
\hat{g}_i = \sum_{j=1}^r \alpha_j \beta_j(\lambda_i),
\end{equation}
where $\beta_1(\lambda), \hdots, \beta_r(\lambda)$ are some fixed interpolation kernels, and $\boldsymbol{\alpha} = (\alpha_1, \hdots, \alpha_r)$ are the interpolation coefficients. In matrix notation, the filter is expressed as 
$\mbox{diag}(\hat{\mathbf{G}}) = \mathbf{B} \boldsymbol{\alpha}$, 
where $\mathbf{B} = (b_{ij}) = (\beta_j(\lambda_i))$ is a $k \times r$ 
matrix. 
%
Such a parametrization results in filters with a number of parameters constant in the input size $n$. 


\paragraph*{Chebyshev Spectral CNN (ChebNet).} In order to alleviate the cost of explicitly computing the graph Fourier transform, Defferrard \etal \cite{defferrard2016convolutional}  used an explicit expansion in the Chebyshev polynomial basis to represent the spectral filters\vspace{-0.5mm} 
\begin{equation} \label{eq:filt_cheby}
	g_{\boldsymbol{\alpha}}(\boldsymbol{\Delta}) = \sum_{j=0}^{r-1} \alpha_j T_j(\tilde{\boldsymbol{\Delta}}) = \sum_{j=0}^{r-1} \alpha_j  \boldsymbol{\Phi} T_j(\tilde{\boldsymbol{\Lambda}})\boldsymbol{\Phi}^\top,\vspace{-0.5mm}
\end{equation}
where $\tilde{\boldsymbol{\Delta}} = 2 \lambda_{n}^{-1}\boldsymbol{\Delta}  - \mathbf{I}$ is the rescaled Laplacian such that its eigenvalues $\tilde{\boldsymbol{\Lambda}} = 2 \lambda_{n}^{-1} \boldsymbol{\Lambda}  - \mathbf{I}$ are in the interval $[-1,1]$, 
$\boldsymbol{\alpha}$ is the $r$-dimensional vector of polynomial coefficients parametrizing the filter, and \vspace{-0.5mm}
\begin{eqnarray}
\label{eq:cheby}
T_j(\lambda) &=& 2\lambda T_{j-1}(\lambda) - T_{j-2}(\lambda), \vspace{-0.5mm}
\end{eqnarray}
denotes the Chebyshev polynomial of degree $j$ defined in a recursive manner with $T_1(\lambda) =\lambda$ and $T_0(\lambda) =1$.

Such an approach has several important advantages. 
First, it does not require an explicit computation of the Laplacian eigenvectors. 
Due to the recursive definition of the Chebyshev polynomials, the computation of the filter $g_{\boldsymbol{\alpha}}(\boldsymbol{\Delta}) \mathbf{f}$ entails applying the Laplacian $r$ times, resulting in $\mathcal{O}(rn)$ operations.
Second, since the Laplacian is a local operator affecting only 1-hop neighbors of a vertex and accordingly its $(r-1)$st power affects the $r$-hop neighborhood, the resulting filters are localized.

\paragraph*{Graph convolutional network (GCN).} 
Kipf and Welling \cite{welling2016} considered  the construction  of \cite{defferrard2016convolutional} with $r=2$, which, under the additional assumption of $\lambda_n \approx 2$, and $\alpha = \alpha_0 = -\alpha_1$ yields single-parametric filters of the form 
$
g_\alpha( \mathbf{f} ) = \alpha ( \mathbf{I} + \mathbf{D}^{-1/2} \mathbf{W} \mathbf{D}^{-1/2}) \mathbf{f} 
$. 
Such a filter is numerically unstable since the maximum eigenvalue of the matrix $\mathbf{I} + \mathbf{D}^{-1/2} \mathbf{W} \mathbf{D}^{-1/2}$ is $2$; a renormalization 
\begin{eqnarray}
\label{eq:welling}
g_\alpha( \mathbf{f} ) &=& \alpha \tilde{\mathbf{D}}^{-1/2} \tilde{\mathbf{W}} \tilde{\mathbf{D}}^{-1/2} \mathbf{f}, 
\end{eqnarray}
with $\tilde{\mathbf{W}} = \mathbf{W} + \mathbf{I}$ and $\tilde{\mathbf{D}} = \mathrm{diag}(\sum_{j \neq i} \tilde{w}_{ij})$ is introduced by the authors in order to cure such problem and allow multiple convolutional levels to be casted one after the other. 


\paragraph*{Diffusion CNN (DCNN).} 
A different spatial-domain method was proposed by Atwood and Towsley \cite{atwood2016search}, who considered a diffusion (random walk) process on the graph. The transition probability of a random walk on a graph is given by $\mathbf{P} = \mathbf{D}^{-1}\mathbf{W}$. Different features are produced by applying diffusion of different length (the powers $\mathbf{P}^0, \hdots, \mathbf{P}^{r-1}$), \vspace{-0.5mm}
$$
\mathbf{f}^{\mathrm{out}}_{l,j}  = \xi( w_{lj}\mathbf{P}^{j}\mathbf{f}^{\mathrm{in}}_{l}), 
$$
where the $n\times p$ and $n\times pr$ matrices $\mathbf{F}^{\mathrm{in}} = (\mathbf{f}^{\mathrm{in}}_1, \hdots, \mathbf{f}^{\mathrm{in}}_p)$  and $\mathbf{F}^{\mathrm{out}} = (\mathbf{f}^{\mathrm{out}}_{1,1}, \hdots, \mathbf{f}^{\mathrm{out}}_{p,r})$ represent the $p$- and $pr$-dimensional input and output signals on the vertices of the graph and $\mathbf{W} = (w_{lj})$ is the $p\times r$ matrix of weights.

%% file: dl_manifolds.tex
\section{Deep learning on manifolds}\label{sec:manifold_cnn}

\newenvironment{psmallmatrix}
  {\left(\begin{smallmatrix}}
  {\end{smallmatrix}\right)}

\begin{table*}[!ht]
	\label{tab:rediscovering-CNN}
    \centering
    \caption{
    Several CNN-type geometric deep learning methods on graphs and manifolds can be obtained as a particular setting of the proposed framework with an appropriate choice of the pseudo-coordinates and weight functions in the definition of the patch operator. 
    $x$ denotes the reference point (center of the patch) and $y$ a point within the patch. 
    $\mathbf{x}$ denotes the Euclidean coordinates on a regular grid. 
    $\bar{\alpha}, \bar{\sigma}_\rho, \bar{\sigma}_\theta$ and $\bar{\mathbf{u}}_j, \bar{\theta}_j, j=1,\hdots, J$ denote fixed parameters of the weight functions.
    }
    \begin{tabular}{ llll}
        \toprule
        Method      	& Pseudo-coordinates  & $\mathbf{u}(x,y)$                       		& Weight function $w_j(\mathbf{u}), j=1,\hdots, J$\\ \midrule
        CNN \cite{lecun1998gradient}         	& Local Euclidean		& $\mathbf{x}(x,y) = \mathbf{x}(y)-\mathbf{x}(x)$ 							& 
        $\delta(\mathbf{u} - \bar{\mathbf{u}}_j)$\\
	GCNN \cite{masci2015geodesic}                    	& Local polar geodesic		&  $\rho(x,y), \theta(x,y)$ 					& 
	$\exp({ -\tfrac{1}{2} (\mathbf{u} - \bar{\mathbf{u}}_j)^\top \begin{psmallmatrix}\bar{\sigma}^2_\rho&\\&\bar{\sigma}^2_\theta\end{psmallmatrix}^{-1} (\mathbf{u} - \bar{\mathbf{u}}_j)  })$\\
	ACNN \cite{boscaini2016learning} & 			 Local polar geodesic		&  $\rho(x,y), \theta(x,y)$ 					&  
	$\exp({ -\tfrac{1}{2} \mathbf{u}^\top \mathbf{R}_{\bar{\theta}_j} \begin{psmallmatrix}\bar{\alpha}&\\&1\end{psmallmatrix} \mathbf{R}_{\bar{\theta}_j}^\top \mathbf{u}  })$	\\		  %
         GCN  \cite{welling2016}     			& Vertex degree &  $\text{deg}(x), \text{deg}(y)$			& $\left(1 - |1 - \tfrac{1}{\sqrt{u_1}}|\right) \left(1 - |1 - \tfrac{1}{\sqrt{u_2}}| \right)$ \\                                                             
	 DCNN \cite{atwood2016search}                	&  Transition probability in $r$ hops & $p^0(x,y),\hdots, p^{r-1}(x,y)$         				&	$\mathrm{id}(u_j)$\\
	 \bottomrule
    \end{tabular}
    \vspace{-3mm}
\end{table*}

Let $\mathcal{X}$ be a $d$-dimensional differentiable manifold, possibly with boundary $\partial \mathcal{X}$. Around point $x \in \mathcal{X}$, the manifold is homeomorphic to a $d$-dimensional Euclidean space referred to as the {\em tangent space} and denoted by $T_x \mathcal{X}$. 
An inner product $\langle \cdot, \cdot\rangle_{T_x \mathcal{X}} : T_x \mathcal{X} \times T_x \mathcal{X} \rightarrow \mathbb{R}$ depending smoothly on $x$ is called the {\em Riemannian metric}. 
In the following, we denote by $f : \mathcal{X} \rightarrow \mathbb{R}$ smooth real functions (scalar fields) on the manifold. In shape analysis, 3D shapes are modeled as 2-dimensional manifolds (surfaces), representing the boundaries of 3D volumes. 

%
%
%

\paragraph*{Geodesic CNN (GCNN).}
Masci \etal \cite{masci2015geodesic} introduced a generalization of CNNs on 2-dimensional manifolds, based on the definition of a local charting procedure in geodesic polar coordinates \cite{isc}. Such a construction, named the {\em patch operator} \vspace{-1mm}
$$
(D(x)f)(\rho,\theta) = \int_\mathcal{X} w_{\rho,\theta}(x,y) f(y) dy  
$$
maps the values of the function $f$ at a neighborhood of the point $x\in \mathcal{X}$ into the local polar coordinates $\rho,\theta$. Here $dy$ denotes the area element induced by the Riemannian metric, and $w_{\rho,\theta}(x,y)$ is a weighting function localized around $\rho,\theta$ (see examples in Figure~\ref{fig:polar_plots}). 
$D(x)f$ can be regarded as a {\em patch} on the manifold; the {\em geodesic convolution}\vspace{-0.5mm}
$$
(f \star g)(x) = \hspace{-0.2cm}\max_{\Delta\theta \in [0,2\pi)} \int_{0}^{2\pi} \hspace{-2mm} \int_0^{\rho_{\max}} \hspace{-5mm} g(\rho, \theta +\Delta\theta)(D(x)f)(\rho, \theta)  d\rho d\theta, 
$$
can be thought of as matching a template $g(\rho,\theta)$ with the extracted patch at each point, 
where the maximum is taken over all possible rotations of the template in order to resolve the origin ambiguity in the angular coordinate.
The geodesic convolution is used to define an analogy of a traditional convolutional layer in GCNN, where the templates $g$ are learned. 


\paragraph*{Anisotropic CNN (ACNN).}
Boscaini \etal \cite{boscaini2016learning} considered the {\em anisotropic diffusion} equation on the manifold
\begin{equation}\label{eq:heat}
f_t(x,t) = -\mathrm{div}_\mathcal{X}( \mathbf{A}(x) \nabla_\mathcal{X} f(x,t) )\,,
\end{equation}
where $\nabla_\mathcal{X}$ and $\mathrm{div}_\mathcal{X}$ denote the {\em intrinsic gradient} and {\em divergence}, respectively, 
$f(x,t)$ is the temperature at point $x$ and time $t$, and the {\em conductivity tensor} $\mathbf{A}(x)$ (operating on the gradient vectors in the tangent space $T_x\mathcal{X}$) allows to model heat flow that is position- and direction-dependent. In particular, they used the $2\times 2$ tensor
\begin{equation}
\mathbf{A}_{\alpha\theta}(x) = \mathbf{R}_\theta(x)\begin{pmatrix}\alpha&\\&1\end{pmatrix}\mathbf{R}^\top_\theta(x)\,,
\end{equation}
where matrix $\mathbf{R}_\theta$ is a rotation by $\theta$ in the tangent plane w.r.t. the maximal curvature direction, and the parameter $\alpha>0$ controls the degree of anisotropy (isotropic diffusion is obtained for $\alpha=1$).
Using as initial condition $f(x,0)$ a point source of heat at $x$, the solution to the heat equation \eqref{eq:heat} is given by the {\em anisotropic heat kernel} $h_{\alpha \theta t} (x,y)$, representing the amount of heat that is transferred from point $x$ to point $y$ at time $t$. 
By varying the parameters $\alpha$, $\theta$ and $t$ (controlling respectively the elongation, orientation, and scale of the kernel) one obtains a collection of kernels that can be used as weighting functions in the construction of the patch operator (see examples in Figure~\ref{fig:polar_plots}). This gives rise to an alternative charting to the geodesic patches of GCNN, more robust to geometric noise, and more efficient to compute.

\input{polar_plots.tex}

Both GCNN and ACNN operate in the spatial domain and thus do not suffer from the inherent inability of spectral methods to generalize across different domains. These methods were shown to outperform all the known hand-crafted approaches for finding intrinsic correspondence between deformable shapes \cite{masci2015geodesic,boscaini2016learning}, a notoriously hard problem in computer graphics. 


%% file: polar_plots.tex
\begin{figure*}[ht!]
\begin{minipage}{0.24\textwidth}
	\centering
	\includegraphics[width=0.9\linewidth]{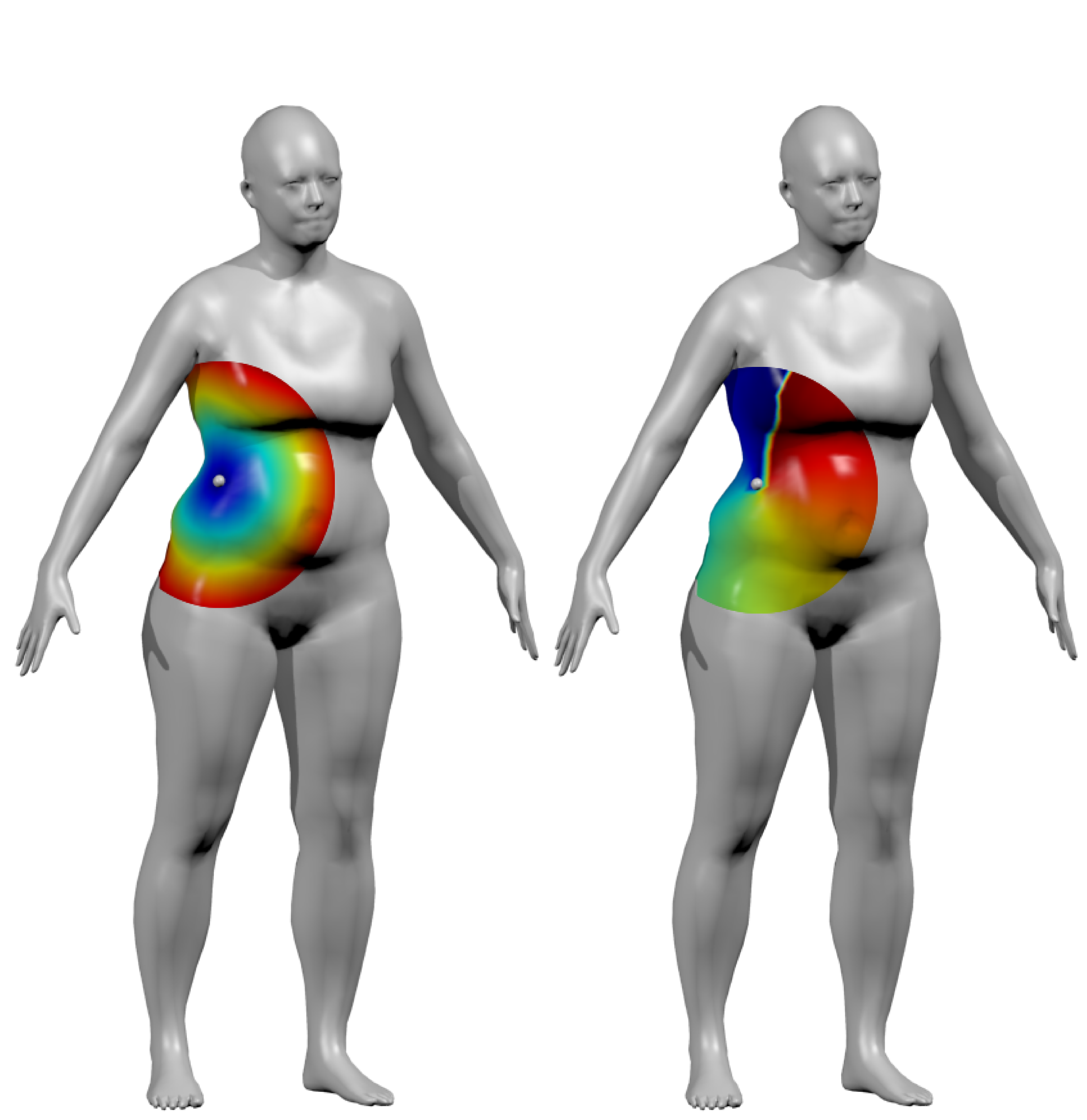}
		\footnotesize Polar coordinates $\rho, \theta$
\end{minipage}
\begin{minipage}{0.24\textwidth}
	\centering
	\setlength\figureheight{4cm} 
	\setlength\figurewidth{4cm} 
	\input{polar_plot_gcnn_kernels}\\
		\footnotesize GCNN
\end{minipage}
\begin{minipage}{0.24\textwidth}
	\centering
	\setlength\figureheight{4cm} 
	\setlength\figurewidth{4cm}
	\input{polar_plot_acnn_kernels}
		\footnotesize ACNN
\end{minipage}
\begin{minipage}{0.24\textwidth}
	\centering
	\setlength\figureheight{4cm} 
	\setlength\figurewidth{4cm}
	\input{polar_plot_monet_kernels_metric}
		\footnotesize MoNet
\end{minipage}\vspace{1mm}
\caption{Left: intrinsic local polar coordinates $\rho, \theta$ on manifold around a point marked in white. 
Right: patch operator weighting functions $w_i(\rho,\theta)$ used in different generalizations of convolution on the manifold (hand-crafted in GCNN and ACNN and learned in MoNet). All kernels are $L_\infty$-normalized; red curves represent the $0.5$ level set. 
}
\vspace{-3.5mm}
\label{fig:polar_plots}
\end{figure*}

%% file: our_method.tex
\section{Our approach}

The main contribution of this paper is a generic spatial-domain framework for deep learning on non-Euclidean domains such as graphs and manifolds. 
%
We use $x$ to denote, depending on context, a point on a manifold or a vertex of a graph, and consider points $y \in \mathcal{N}(x)$ in the neighborhood of $x$. 
With each such $y$, we associate a $d$-dimensional vector of {\em pseudo-coordinates} $\mathbf{u}(x,y)$. 
In these coordinates, we define a weighting function (kernel) $\mathbf{w}_{\boldsymbol{\Theta}}(\mathbf{u}) = (w_1(\mathbf{u}), \hdots, w_J(\mathbf{u}))$, which is parametrized by some learnable parameters $\boldsymbol{\Theta}$. 
%
%
The patch operator can therefore be written in the following general form  
\begin{equation}
\label{eq:our_patch}
D_j(x) f = \sum_{y\in\mathcal{N}(x)} w_j(\mathbf{u}(x,y)) f(y), \,\,\,\,\, j = 1,\hdots, J,
\end{equation}
where the summation should be interpreted as an integral in the case we deal with a continuous manifold, and $J$ represents the dimensionality of the extracted patch. 
%
%
A spatial generalization of the convolution on non-Euclidean domains is then given by a template-matching procedure of the form 
\begin{equation}
\label{eq:our_conv}
(f\star g)(x) = \sum_{j=1}^J g_j \, D_j(x) f. 
\end{equation}

The two key choices in our construction are the pseudo-coordinates $\mathbf{u}$ and the weight functions $\mathbf{w}(\mathbf{u})$. 
Table~\ref{tab:rediscovering-CNN} shows that other deep learning methods (including the classical CNN on Euclidean domains, DCN and DCNN on graphs, and GCNN and ACNN on manifolds) can be obtained as particular settings of our framework with appropriate definition of $\mathbf{u}$ and $\mathbf{w}(\mathbf{u})$. 
For example, GCNN and ACNN boil down to using Gaussian kernels on local polar geodesic coordinates $\rho,\theta$ on a manifold, and GCN can be interpreted as applying a triangular kernel on pseudo-coordinates given by the degree of the graph vertices.

In this paper, rather than using fixed handcrafted weight functions we consider parametric kernels with learnable parameters. 
In particular, a convenient choice is 
\begin{equation}
\label{eq:our_w}
w_j(\mathbf{u}) = \exp(-\tfrac{1}{2}(\mathbf{u}-\boldsymbol{\mu}_j)^\top \boldsymbol{\Sigma}_j^{-1} (\mathbf{u}-\boldsymbol{\mu}_j)), 
\end{equation}
where $\boldsymbol{\Sigma}_j$ and $\boldsymbol{\mu}_j$ are learnable $d\times d$ and $d\times 1$ covariance matrix and mean vector of a Gaussian kernel, respectively. 
Formulae~(\ref{eq:our_patch}--\ref{eq:our_conv}) can thus be interpreted as a gaussian mixture model (GMM). 
We further restrict the covariances to have diagonal form, resulting in $2d$ parameters per kernel, and a total of $2Jd$ parameters for the patch operator.

While extremely simple, we show in the next section that these additional degrees of freedom afford our architecture sufficient complexity allowing it to outperform existing approaches. 
More complex versions of the weighting functions could include additional non-linear transformation of the pseudo-coordinates $\mathbf{u}$ before feeding them to the Gaussian kernel, or even more general network-in-a-network architectures \cite{LinCY13}.

%% file: results_images.tex
\subsection{Images}

\begin{figure}
\begin{minipage}[t]{0.46\linewidth}
\includegraphics[width=\linewidth]{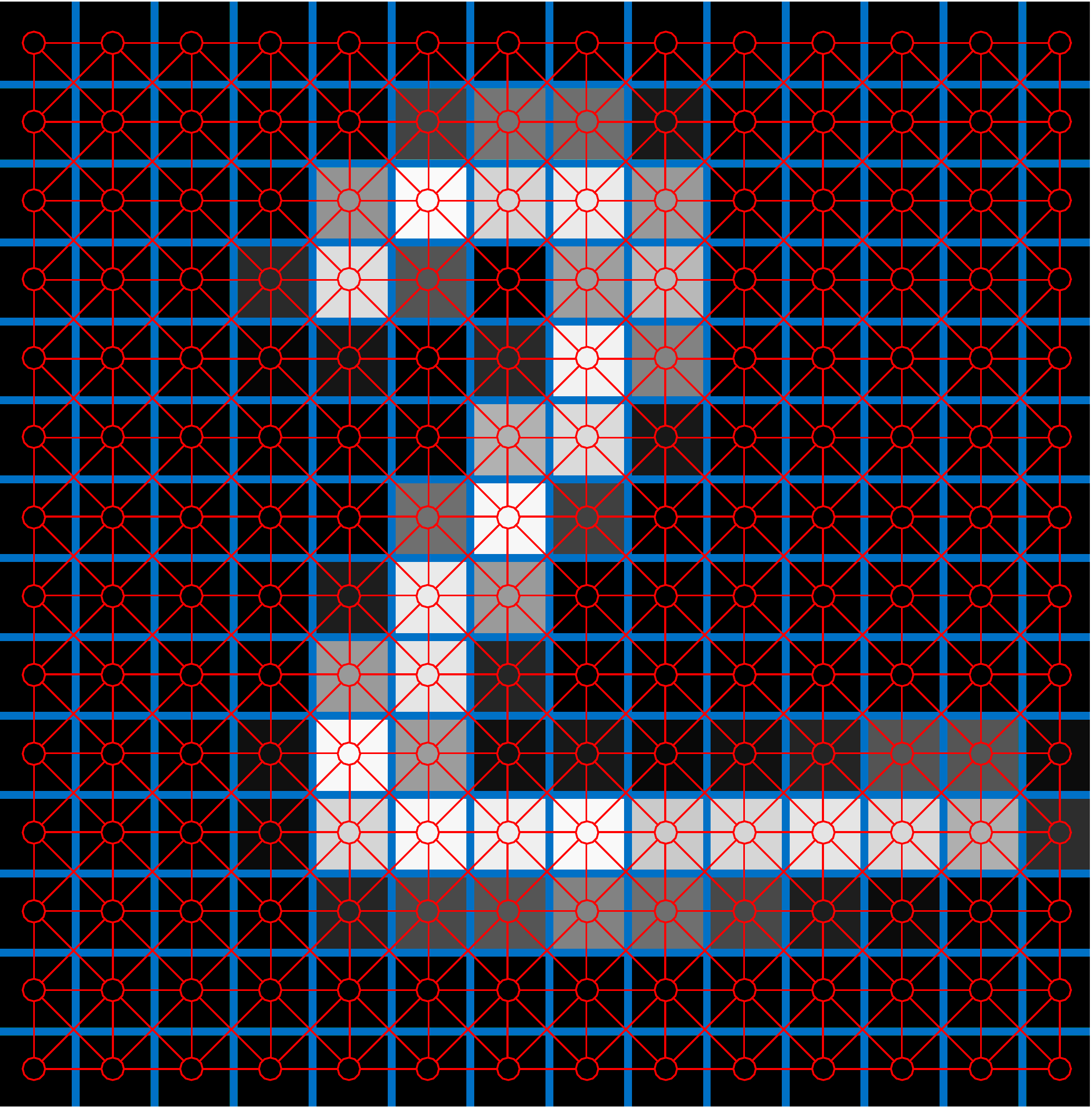}
\end{minipage}\hspace{3mm}
\begin{minipage}[t]{0.46\linewidth}
\includegraphics[width=\linewidth]{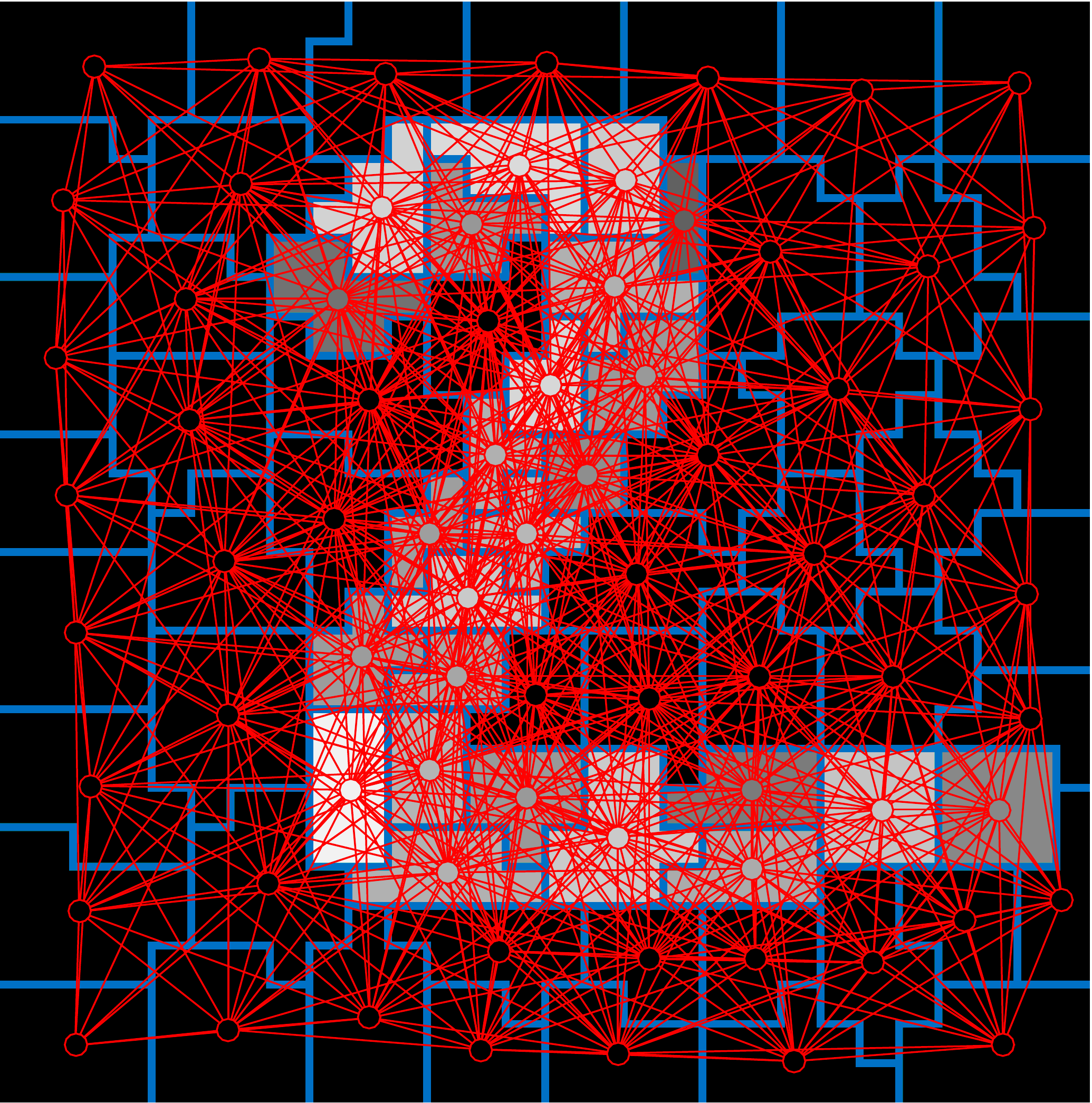}
\end{minipage}\vspace{3mm}\\
\begin{minipage}[t]{0.46\linewidth}
\includegraphics[width=\linewidth]{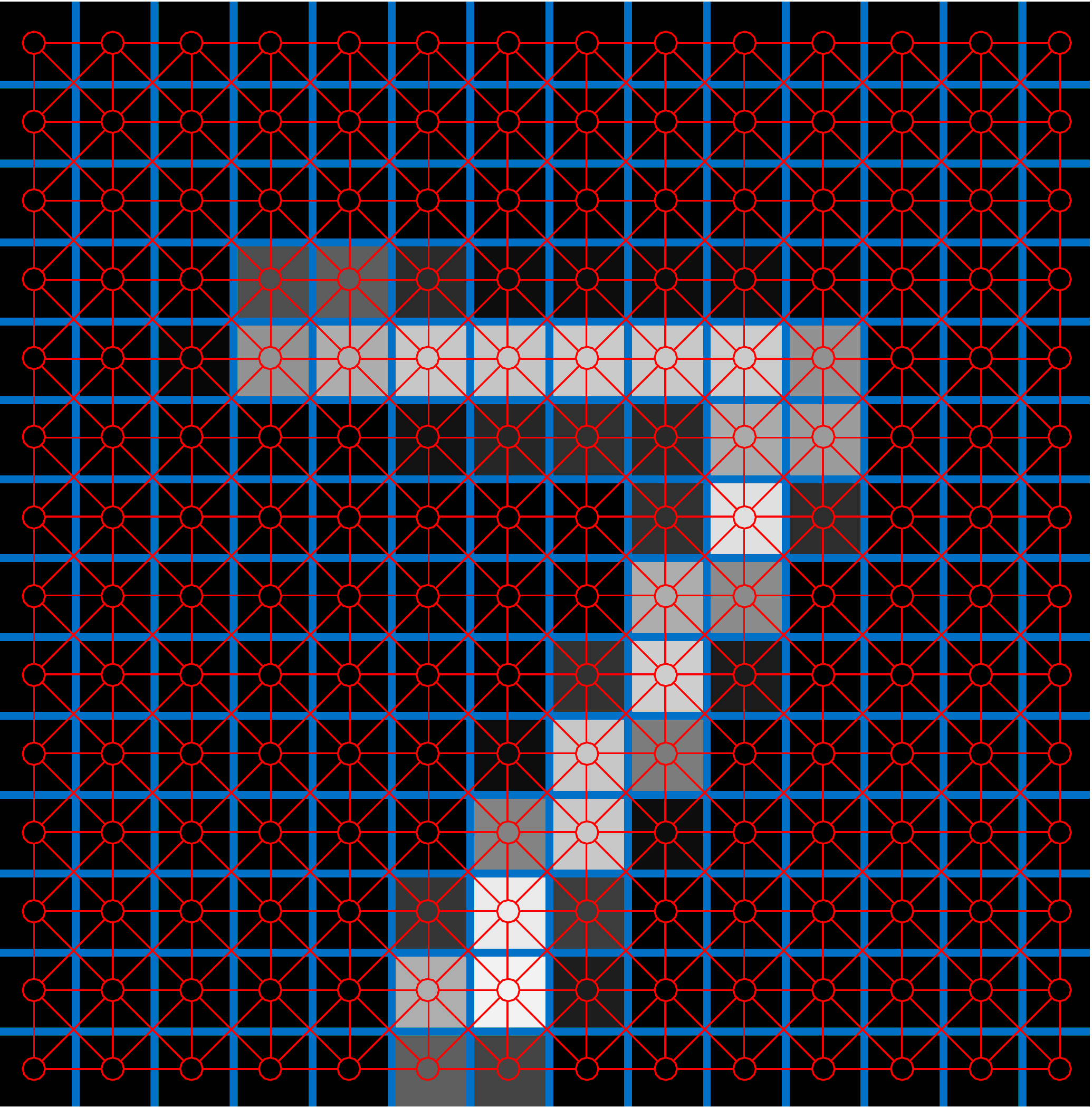}
\end{minipage}\hspace{3mm}
\begin{minipage}[t]{0.46\linewidth}
\includegraphics[width=\linewidth]{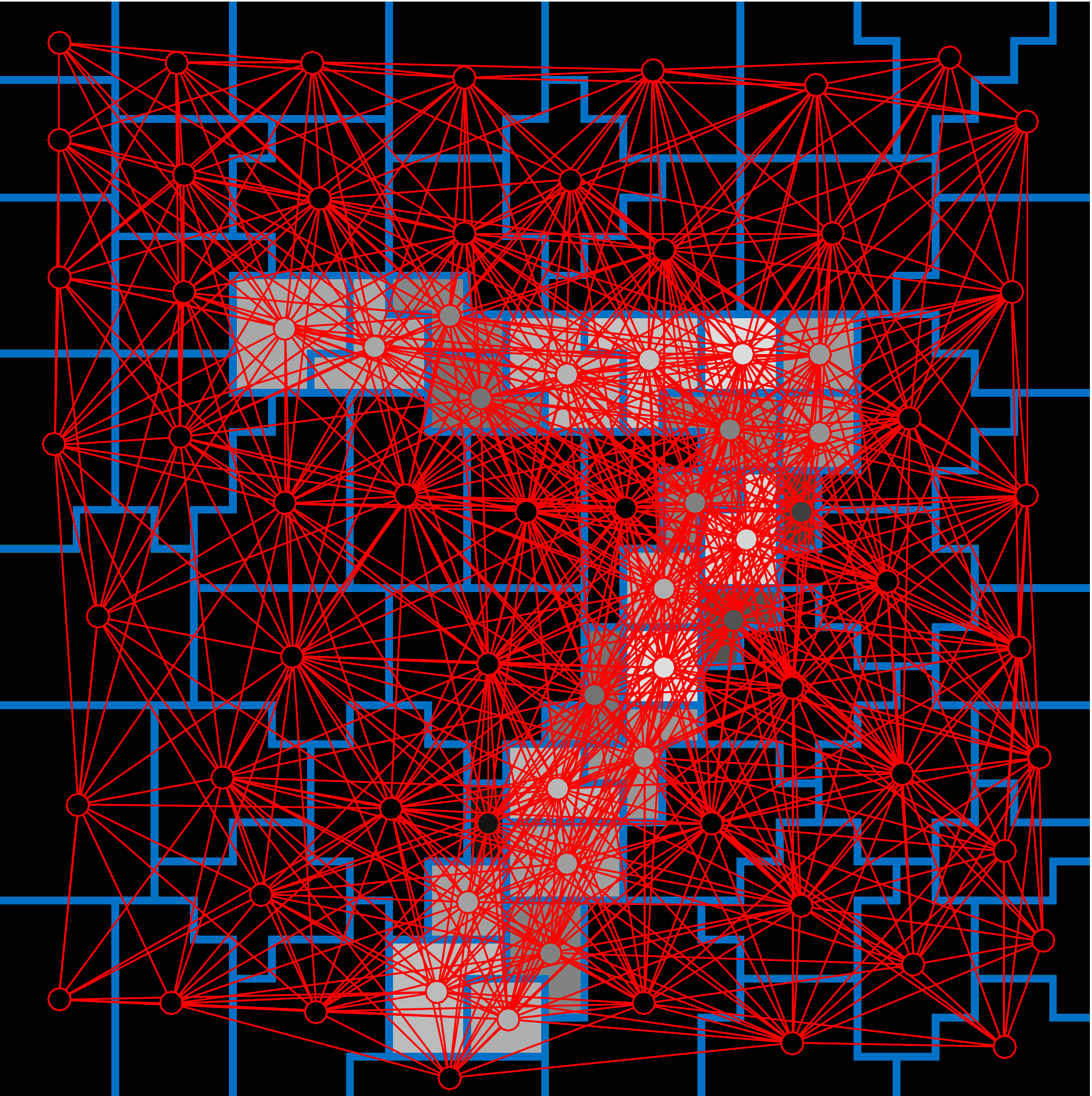}
\end{minipage}\\
\begin{minipage}[t]{0.46\linewidth}
\centering\small Regular grid
\end{minipage}\hspace{3mm}
\begin{minipage}[t]{0.46\linewidth}
\centering\small Superpixels
\end{minipage}\vspace{0.5mm}
\caption{
Representation of images as graphs. Left: regular grid (the graph is fixed for all images). Right: graph of superpixel adjacency (different for each image). Vertices are shown as red circles, edges as red lines. 
\label{fig:superpixels}
}
\vspace{-2.5mm}
\end{figure}

In our first experiment, we applied the proposed method on a classical task of handwritten digit classification in the MNIST dataset \cite{lecun1998gradient}. While almost trivial by todays standards, we nevertheless use this example to visualize an important advantage of our approach over spectral graph CNN methods. 
Our experimental setup followed \cite{defferrard2016convolutional}. 
The $28\times 28$ images were represented as graphs, where vertices correspond to (super)pixels and edges represent their spatial relations. We considered two constructions: all images represented on the {\em same graph} (regular grid) and each image represented as a {\em different graph} (Figure~\ref{fig:superpixels} left and right, respectively). 
Furthermore, we varied the graph size: the full and $\tfrac{1}{4}$ grids contained $728$ and $196$ vertices, respectively, while the superpixel-based graphs contained $300$, $150$, and $75$ vertices.

Three methods were compared: classical CNN LeNet5 architecture \cite{lecun1998gradient} (containing two convolutional, two max-pooling, and one fully-connected layer, applied on regular grids only), spectral ChebNet\cite{defferrard2016convolutional} and the proposed MoNet. We used a standard splitting of the MNIST dataset into training-, testing-, and validation sets of sizes 55K, 10K, and 5K images, respectively. 
LeNet used $2\times 2$ max-pooling; in ChebNet and MoNet we used three convolutional layers, interleaved with pooling layers based on the Graclus method \cite{dhillon2007weighted} to coarsen the graph by a factor of four.

For MoNet, we used polar coordinates $\mathbf{u} = (\rho, \theta)$ of pixels (respectively, of superpixel barycenters) to produce the patch operator; as the weighting functions of the patch operator, $25$ Gaussian kernels (initialized with random means and variances) were used. 
Training was done with 350K iterations of Adam method \cite{KingmaB14}, initial learning rate $10^{-4}$, 
regularization factor $10^{-4}$, dropout probability $0.5$, and batch size of $10$.

Table~\ref{tab:mnist} summarizes the performance of different algorithms. 
On regular grids, all the methods perform approximately equally well. 
However, when applying ChebNet on superpixel-based representations, the performance drops dramatically (by up to almost 25\%). The reason lies in the key drawback of spectral CNN models, wherein the definition of the filters is basis- and thus domain-dependent. Since in this case each image is represented as a different graph, the model fails to generalize well. The effect is most pronounced on smaller graphs (150 and 75 superpixels) that vary strongly among each other. In contrast, the proposed MoNet approach manifests consistently  high accuracy, and only a light performance degradation is observed when the image presentation is too coarse (75 superpixels).


%
%
%
%

\begin{table}[H]
\centering
\caption{Classification accuracy of classical Euclidean CNN (LeNet5), spectral CNN (ChebNet) and the proposed approach (MoNet) on different versions of the MNIST dataset.  
The setting of all the input images sharing the same graph is marked with *. 
\vspace{2mm}
}
\label{tab:mnist}
\begin{tabular}{@{}lccc@{}}
\toprule
Dataset               & LeNet5 \cite{lecun1998gradient} & \multicolumn{1}{l}{ChebNet \cite{defferrard2016convolutional}} & \multicolumn{1}{l}{\textbf{MoNet}} \\ \midrule
*Full grid                & 99.33\%  & 99.14\%                       & {99.19\%}                            \\
*$\tfrac{1}{4}$ grid             & 98.59\%      & 97.70\%                        & {98.16\%}                           \\
\hline 
300 Superpixels & -      & 88.05\%                       & \textbf{97.30\%}                            \\
150 Superpixels & -      & 80.94\%                       & \textbf{96.75\%}                            \\
75 Superpixels  & -      & 75.62\%                       & \textbf{91.11\%}                            \\ \bottomrule
\end{tabular}
\vspace{-0.5mm}
\end{table}

%% file: results_graphs.tex
\subsection{Graphs}

In the second experiment, we address the problem of vertex classification on generic graphs. 
%
We used the popular Cora and PubMed \cite{aimag08} citation graphs as our datasets. In each dataset, a vertex
represents a scientific publication (2708 vertices in Cora and 19717 in PubMed, respectively), and an undirected unweighted edge represents a citation (5429 and 44338 edges in Cora and PubMed). For each vertex, a feature vector representing the content of the paper is given (1433-dimensional binary feature vectors in Cora, and 500-dimensional tf-idf weighted word vectors in PubMed). The task is to classify each vertex into one of the groundtruth classes (7 in Cora and 3 in PubMed). 
%

We followed verbatim the experimental settings presented in \cite{yang2016revisiting, welling2016}. The training sets consisted of 20 samples per class; the validation and test sets consisted of 500 and 1000 disjoint vertices. 
The validation set was chosen in order to reflect the probability distribution of the various classes over the entire dataset. 
We compared our approach to all the methods compared in \cite{welling2016}.

For MoNet, we used the degrees of the nodes as the input pseudo-coordinates $\mathbf{u}(x,y) = (\tfrac{1}{\sqrt{\mathrm{deg(x)}}}, \tfrac{1}{\sqrt{\mathrm{deg(y)}}})^\top$; these coordinates underwent an additional transformation in the form of a fully-connected neural network layer 
$
\tilde{\mathbf{u}}(x,y) = \mathrm{tanh}(\mathbf{A}\mathbf{u}(x,y) + \mathbf{b})
$, 
where the $r\times 2$ matrix $\mathbf{A}$ and $r\times 1$ vector $\mathbf{b}$ were also learned (we used $r=2$ for Cora and $r=3$ for PubMed). 
The Gaussian kernels were applied on coordinates $\tilde{\mathbf{u}}(x,y)$ yielding patch operators of the form
$$
D_j(x) f_l = \sum_{y \in \mathcal{N}(x)} e^{-\tfrac{1}{2}(\tilde{\mathbf{u}}(x,y) - \boldsymbol{\mu}_j)^\top\boldsymbol{\Sigma}^{-1}_j (\tilde{\mathbf{u}}(x,y) - \boldsymbol{\mu}_j)} f_l(y),
$$
where $\boldsymbol{\Sigma}_j$, $\boldsymbol{\mu}_j$, $j=1,\hdots, J$ are the $r\times r$ and $r\times 1$ covariance matrices and mean vectors of the Gaussian kernels, respectively. 
DCNN, GCN and MoNet were trained in the same way in order to give a fair comparison (see training details in Table~\ref{tab:learning-conf}). 
The $L_2$-regularization weights for MoNet were $\gamma = 10^{-2}$ and $5\times 10^{-2}$ for Cora and PubMed, respectively;  for DCNN and GCN we used the values suggested by the authors in \cite{atwood2016search} and \cite{welling2016}.

\begin{table}[]
\centering
\caption{Learning configuration used for Cora and PubMed experiments.}
\label{tab:learning-conf}
\begin{tabular}{@{}lll@{}}
\toprule
\multicolumn{1}{c}{}                     & \multicolumn{1}{l}{Cora} & \multicolumn{1}{l}{PubMed} \\ \midrule
Learning Algorithm                       & Adam                     & Adam                       \\
Number of epochs                         & 3000                     & 1000                       \\
Validation frequency        		&0.01                     & 0.04                         \\
Learning rate                            & 0.1                      & 0.1                        \\
Decay rate                                    & $10^{-1}$                & -                          \\
Decay epochs                             & 1500, 2500               & -                          \\
Early stopping                           & No                       & No                         \\ \bottomrule
\end{tabular}
\end{table}

\begin{figure*}[h!t]
\centering
\begin{minipage}[t]{1\linewidth}
\includegraphics[width=0.95\linewidth]{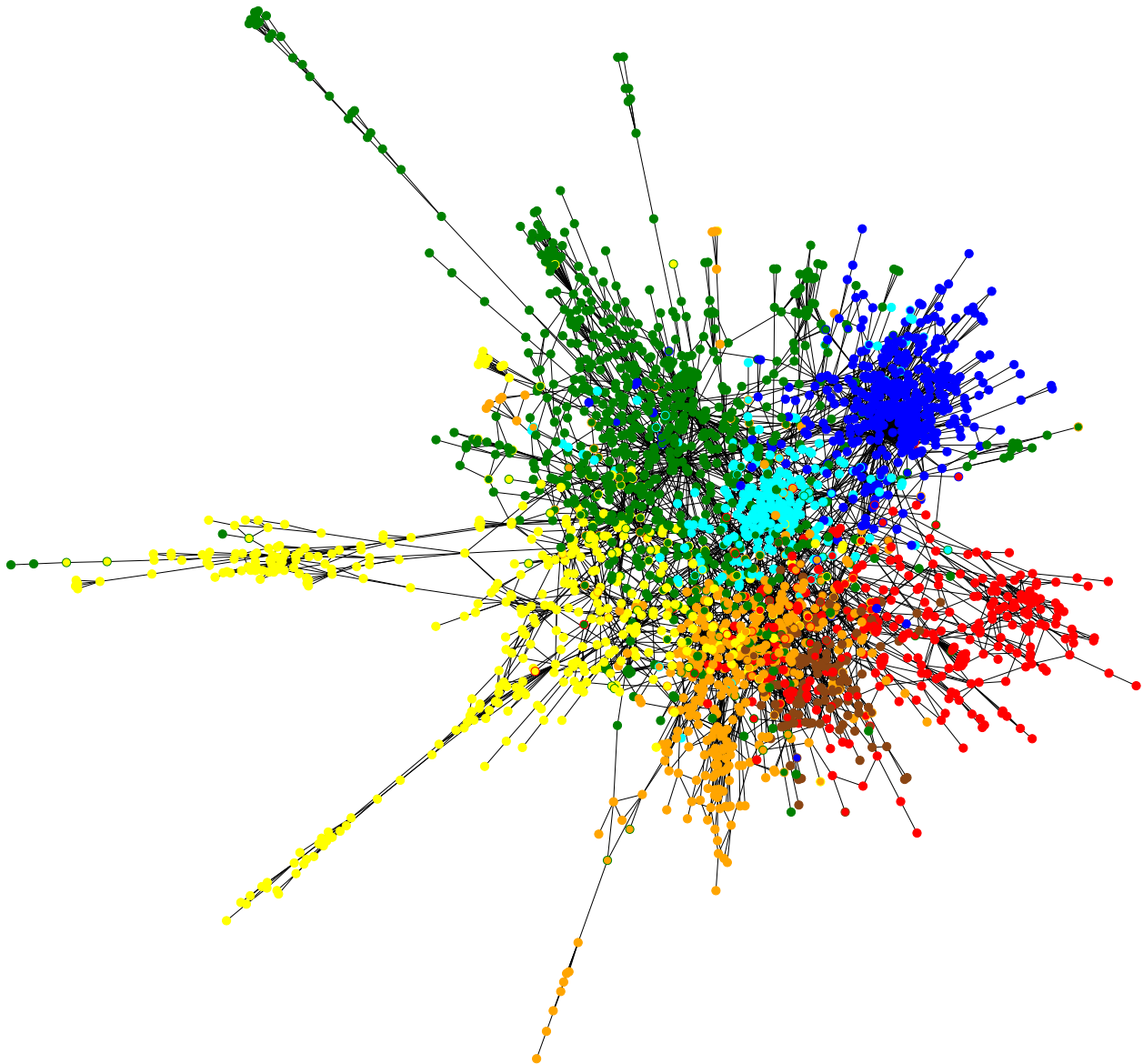}
\end{minipage}\vspace{2mm}
\caption{Predictions obtained applying MoNet over the Cora dataset. Marker fill color represents the predicted class; marker outline color represents the groundtruth class. }
\label{fig:cora-pred}
\end{figure*}

The vertex classification results of different methods are summarized in Table~\ref{tab:planetoid-splitting-res} and visualized in Figure~\ref{fig:cora-pred}. 
MoNet compares favorably to other approaches. 
The tuning of the network hyper-parameters has been fundamental in this case for avoiding overfitting, due to a very small size of the training set. Being more general, our architecture is more complex compared to GCN and  DCNN and requires an appropriate regularization to be used in such settings. 
At the same time, the greater complexity of our framework might prove advantageous when applied to larger and more complex data.

\begin{table}
\centering
\caption{Vertex classification accuracy on the Cora and PubMed datasets following the splitting suggested in \cite{yang2016revisiting}. Learning methods (DCNN, GCNN and MoNet) were trained and tested fifty times for showing their average behavior with different initializations.}
\label{tab:planetoid-splitting-res}
\begin{tabular}{@{}lll@{}}
\toprule
Method & Cora   & PubMed \\ \midrule
ManiReg \cite{belkin2006manifold}  & 59.5\%   & 70.7\%    \\
SemiEmb \cite{weston2012deep}  & 59.0\%     & 71.1\%  \\
LP \cite{zhu2003semi}       & 68.0\%   & 63.0\%    \\
DeepWalk \cite{perozzi2014deepwalk}  & 67.2\%    & 65.3\%   \\
Planetoid \cite{yang2016revisiting} & 75.7\%    & 77.2\%   \\ \midrule
DCNN \cite{atwood2016search}     & 76.80 $\pm$ 0.60\% & 73.00  $\pm$ 0.52\% \\
GCN \cite{welling2016}        & 81.59 $\pm$ 0.42\% & 78.72  $\pm$ 0.25\% \\
\textbf{MoNet}     & \textbf{81.69} $\pm$ 0.48\% & \textbf{78.81}  $\pm$ 0.44\%\\ \bottomrule
\end{tabular}
\vspace{-2mm}
\end{table}

%% file: results_shapes.tex
\subsection{Manifolds}

The last application we consider is learning dense intrinsic correspondence between collections of 3D shapes represented as discrete manifolds. 
For this purpose, correspondence is cast as a labelling problem, where one tries to label each vertex of a given query shape $\mathcal{X}$ with the index of a corresponding point on some reference shape $\mathcal{Y}$ \cite{rodola2014dense,masci2015geodesic,boscaini2016learning}. Let $n$ and $m$ denote the number of vertices in $\mathcal{X}$ and $\mathcal{Y}$, respectively. 
For a point $x$ on a query shape, the last layer of the network is soft-max, producing an $m$-dimensional output $\mathbf{f}
(x)$ that is interpreted as a probability distribution on $\mathcal{Y}$ (the probability of $x$ mapped to $y$). 
Learning is done by minimizing the standard logistic regression cost \cite{boscaini2016learning}. 

\paragraph{Meshes.} 
We reproduced verbatim the experiments of \cite{masci2015geodesic,boscaini2016learning} on the FAUST humans dataset~\cite{bogo}, comparing to the methods reported therein. 
The dataset consisted of $100$ watertight meshes representing $10$ different poses for $10$ different subjects with exact ground-truth correspondence. 
Each shape was represented as a mesh with 6890 vertices; the first subject in first pose was used as the reference. 
For all the shapes, point-wise 544-dimensional SHOT descriptors (local histogram of normal vectors) \cite{tombari2010unique} were used as input data. 
We used MoNet architecture with 3 convolutional layers, replicating the architectures of \cite{masci2015geodesic,boscaini2016learning}. 
First 80 subjects in all the poses were used for training (800 shapes in total); the remaining 20 subjects were used for testing. 
The output of the network was refined using the intrinsic Bayesian filter \cite{bayesian} in order to remove some local outliers.

Correspondence quality was evaluated using the Princeton benchmark \cite{kim2011blended}, 
plotting the percentage of matches that are at most $r$-geodesically distant from the groundtruth correspondence on the reference shape. 
For comparison, we report the performance of blended maps \cite{kim2011blended}, random forests \cite{rodola2014dense}, GCNN \cite{masci2015geodesic}, ADD \cite{boscaini2016anisotropic}, and ACNN \cite{boscaini2016learning}. 


%

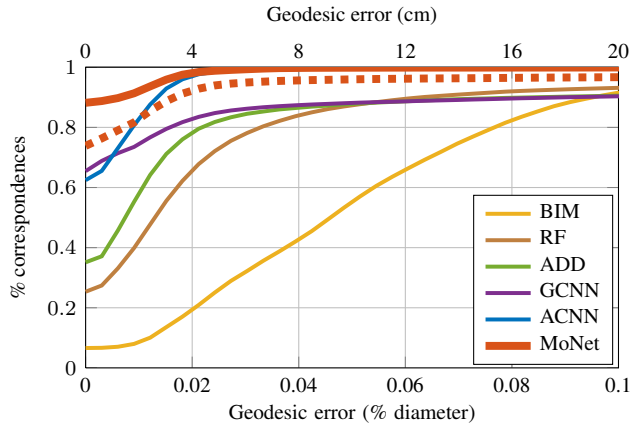
\begin{figure}[h!]
\vspace{-2.5mm}
\begin{minipage}{1.0\linewidth}
	\centering
	\setlength\figureheight{4cm} 
	\setlength\figurewidth{\linewidth}
	\input{Kim_comparison_final.tikz}
\end{minipage}\vspace{-2.5mm}
\caption{Shape correspondence quality obtained by different methods on the FAUST humans dataset. The raw performance of MoNet is shown in dotted curve.} 
\label{fig:kim_curves_raw}\vspace{-1.5mm}
\end{figure}

Figure~\ref{fig:polar_plots} shows the weighting functions of the patch operator that are fixed in GCNN and ACNN architectures, and part of the learnable parameters in the proposed MoNet. The patch operators of GCNN and ACNN can be obtained as a particular configuration of MoNet, implying that if trained correctly, the new model can only improve w.r.t. the previous ones. 
Figure~\ref{fig:kim_curves_raw} depicts the evaluation results, showing that MoNet significantly outperforms the competing approaches. In particular, close to 90\% of points have {\em zero} error, and for 99\% of the points the error is below 4cm. 
%
Figure~\ref{fig:geod_errs} shows the point-wise geodesic correspondence error of our method, and  
Figure~\ref{fig:corrs_as_rgb} visualizes the obtained correspondence using texture transfer.

\begin{figure}[h!]
\vspace{-3mm}
\begin{minipage}{1.0\linewidth}
	\centering
	\setlength\figureheight{4cm} 
	\setlength\figurewidth{\linewidth}
	\input{Kim_comparison_range_scans.tikz}
\end{minipage}\vspace{-2.5mm}
\caption{Shape correspondence quality obtained by different methods on FAUST range maps. 
For comparison, we show the performance of a Euclidean CNN with a comparable 3-layer architecture. 
The raw performance is shown as dotted curve.}
\label{fig:kim_curves_range_scans}\vspace{-1.5mm}
\vspace*{-\baselineskip}
\end{figure}
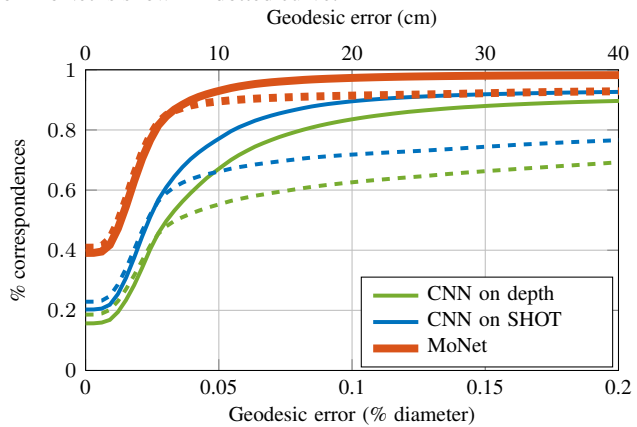

\input{geod_errs_all.tex}

\input{corrs_as_rgb.tex}

\paragraph{Range maps. } Finally, we repeated the shape correspondence experiment on range maps synthetically generated from FAUST meshes. For each subject and pose, we produced 10 rangemaps in 100$\times$180 resolution, covering shape rotations around the $z$-axis with increments of $36$ degrees (total of 1000 range maps), keeping the groundtruth correspondence.    
We used MoNet architecture with 3 convolutional layers and local SHOT descriptors as input data. 
Training and testing set splitting was done as previously.

Figure~\ref{fig:kim_curves_range_scans} shows the quality of correspondence computed using the Princeton protocol. For comparison, we show the performance of a standard Euclidean CNN in equivalent architecture (3 convolutional layers) applied on raw depth values and on SHOT descriptors. Our approach clearly shows a superior performance. 
Figure~\ref{fig:geod_errs_range_maps} shows the point-wise geodesic correspondence error. 
Figure~\ref{fig:corrs_as_rgb_range_scans} shows a qualitative visualization of correspondence using similar color code for corresponding vertices. We also show correspondence on shapes from SCAPE \cite{anguelov2005scape} and TOSCA \cite{bronstein2008numerical} datasets. 


\input{geod_errs_range_scans_all.tex}

\input{corrs_as_rgb_range_scans.tex}

%% file: Kim_comparison_final.tikz
%
%
\definecolor{cool_yellow}{rgb}{0.92900,0.69400,0.12500}%
\definecolor{cool_green}{rgb}{0.46600,0.67400,0.18800}%
\definecolor{cool_purple}{rgb}{0.49400,0.18400,0.55600}%
\definecolor{cool_blue}{rgb}{0.00000,0.44700,0.74100}%
\definecolor{cool_red}{rgb}{0.85000,0.32500,0.09800}%
\begin{tikzpicture}

\pgfplotsset{compat=newest} 

\tikzstyle{every node}=[font=\footnotesize]

\begin{axis}[%
width=0.85\figurewidth,
height=\figureheight,
scale only axis,
xmin=0,
xmax=0.2,
xlabel={Geodesic error (cm)},
xtick = {0,0.04,0.08,0.12,0.16,0.2},
xticklabels = {0,4,8,12,16,20},
axis x line*=top,
axis y line=none,
]
\addplot [color=white,opacity=0.0,line width=1pt,forget plot]
  table[row sep=crcr]{%
0	0.0661103047895501\\
0.203030303030303	0.998846153846154\\
};
\end{axis}

\begin{axis}[%
width=0.85\figurewidth,
height=\figureheight,
at={(0\figurewidth,0\figureheight)},
scale only axis,
xmin=0,
xmax=0.1,
xlabel style={align=center,text width=5cm},
xlabel={\vspace{-0.5cm}\begin{center}Geodesic error (\% diameter) 
\end{center}},
xtick={0,0.02,0.04,0.06,0.08,0.1},
xticklabels = {0,0.02,0.04,0.06,0.08,0.1},
xmajorgrids,
ymin=0,
ymax=1,
ytick={0,0.2,0.4,0.6,0.8,1},
ylabel={\% correspondences},
ymajorgrids,
axis background/.style={fill=white},
legend style={
	at={(0.99,0.01)},
	anchor=south east,
	legend cell align=left,
	align=left,
	text width=2.25em,
	text height=1ex,
}
]

\addplot [color=cool_yellow,solid,line width=1.5pt]
  table[row sep=crcr]{%
0	0.0661103047895501\\
0.00303030303030303	0.066799709724238\\
0.00606060606060606	0.070355587808418\\
0.00909090909090909	0.0798185776487663\\
0.0121212121212121	0.100573294629898\\
0.0151515151515152	0.135609579100145\\
0.0181818181818182	0.1705587808418\\
0.0212121212121212	0.209150943396226\\
0.0242424242424242	0.250674891146589\\
0.0272727272727273	0.289419448476052\\
0.0303030303030303	0.322184325108853\\
0.0333333333333333	0.35621915820029\\
0.0363636363636364	0.387626995645864\\
0.0393939393939394	0.420471698113208\\
0.0424242424242424	0.454818577648766\\
0.0454545454545455	0.493730043541364\\
0.0484848484848485	0.532953555878084\\
0.0515151515151515	0.570464441219158\\
0.0545454545454545	0.605355587808418\\
0.0575757575757576	0.635979680696662\\
0.0606060606060606	0.664499274310595\\
0.0636363636363636	0.692300435413643\\
0.0666666666666667	0.718534107402032\\
0.0696969696969697	0.745616835994194\\
0.0727272727272727	0.769637155297533\\
0.0757575757575758	0.79288824383164\\
0.0787878787878788	0.815566037735849\\
0.0818181818181818	0.835275761973875\\
0.0848484848484849	0.853200290275762\\
0.0878787878787879	0.870377358490566\\
0.0909090909090909	0.884056603773585\\
0.0939393939393939	0.895478955007257\\
0.096969696969697	0.906088534107402\\
0.1	0.915507982583454\\
0.103030303030303	0.923200290275762\\
0.106060606060606	0.930384615384616\\
0.109090909090909	0.937314949201742\\
0.112121212121212	0.944267053701016\\
0.115151515151515	0.950631349782293\\
0.118181818181818	0.956095791001451\\
0.121212121212121	0.960907111756169\\
0.124242424242424	0.96589259796807\\
0.127272727272727	0.970341074020319\\
0.13030303030303	0.974484760522496\\
0.133333333333333	0.97900580551524\\
0.136363636363636	0.982714078374456\\
0.139393939393939	0.986081277213353\\
0.142424242424242	0.989049346879536\\
0.145454545454545	0.99166908563135\\
0.148484848484848	0.993584905660377\\
0.151515151515152	0.995297532656023\\
0.154545454545455	0.996487663280116\\
0.157575757575758	0.99711175616836\\
0.160606060606061	0.997496371552975\\
0.163636363636364	0.997917271407837\\
0.166666666666667	0.998272859216255\\
0.16969696969697	0.99855587808418\\
0.172727272727273	0.998744557329463\\
0.175757575757576	0.998838896952104\\
0.178787878787879	0.998846153846154\\
0.181818181818182	0.998846153846154\\
0.184848484848485	0.998846153846154\\
0.187878787878788	0.998846153846154\\
0.190909090909091	0.998846153846154\\
0.193939393939394	0.998846153846154\\
0.196969696969697	0.998846153846154\\
0.2	0.998846153846154\\
0.203030303030303	0.998846153846154\\
};
\addlegendentry{BIM};
\addplot [color=brown,solid,line width=1.5pt]
  table[row sep=crcr]{%
0	0.253345428156749\\
0.00303030303030303	0.274100145137881\\
0.00606060606060606	0.331262699564586\\
0.00909090909090909	0.398715529753266\\
0.0121212121212121	0.477358490566038\\
0.0151515151515152	0.555645863570392\\
0.0181818181818182	0.622249637155298\\
0.0212121212121212	0.677866473149492\\
0.0242424242424242	0.721088534107402\\
0.0272727272727273	0.755312046444122\\
0.0303030303030303	0.781342525399129\\
0.0333333333333333	0.803483309143686\\
0.0363636363636364	0.821473149492017\\
0.0393939393939394	0.837191582002903\\
0.0424242424242424	0.849956458635704\\
0.0454545454545455	0.860936139332366\\
0.0484848484848485	0.869673439767779\\
0.0515151515151515	0.877140783744557\\
0.0545454545454545	0.883875181422351\\
0.0575757575757576	0.890362844702468\\
0.0606060606060606	0.896255442670537\\
0.0636363636363636	0.901161103047895\\
0.0666666666666667	0.904992743105951\\
0.0696969696969697	0.908730043541364\\
0.0727272727272727	0.912155297532656\\
0.0757575757575758	0.915500725689405\\
0.0787878787878788	0.918860667634253\\
0.0818181818181818	0.92133526850508\\
0.0848484848484849	0.923309143686502\\
0.0878787878787879	0.925174165457184\\
0.0909090909090909	0.926959361393324\\
0.0939393939393939	0.928577648766328\\
0.096969696969697	0.930036284470247\\
0.1	0.931574746008708\\
0.103030303030303	0.932721335268505\\
0.106060606060606	0.933853410740203\\
0.109090909090909	0.934912917271408\\
0.112121212121212	0.936030478955007\\
0.115151515151515	0.937097242380261\\
0.118181818181818	0.937939042089985\\
0.121212121212121	0.938976777939042\\
0.124242424242424	0.939658925979681\\
0.127272727272727	0.940246734397678\\
0.13030303030303	0.940950653120464\\
0.133333333333333	0.941516690856314\\
0.136363636363636	0.942039187227867\\
0.139393939393939	0.942656023222061\\
0.142424242424242	0.94310595065312\\
0.145454545454545	0.943606676342526\\
0.148484848484848	0.944063860667634\\
0.151515151515152	0.944506531204644\\
0.154545454545455	0.944992743105951\\
0.157575757575758	0.94533381712627\\
0.160606060606061	0.945732946298984\\
0.163636363636364	0.946146589259797\\
0.166666666666667	0.946589259796807\\
0.16969696969697	0.947010159651669\\
0.172727272727273	0.947416545718433\\
0.175757575757576	0.947866473149492\\
0.178787878787879	0.948345428156749\\
0.181818181818182	0.948904208998549\\
0.184848484848485	0.949375907111756\\
0.187878787878788	0.949927431059507\\
0.190909090909091	0.950566037735849\\
0.193939393939394	0.9511175616836\\
0.196969696969697	0.951640058055153\\
0.2	0.952184325108854\\
0.203030303030303	0.952924528301887\\
};
\addlegendentry{RF};

\addplot [color=cool_green,solid,line width=1.5pt]
  table[row sep=crcr]{%
0	0.351168359941945\\
0.00303030303030303	0.371277213352685\\
0.00606060606060606	0.457227866473149\\
0.00909090909090909	0.552198838896952\\
0.0121212121212121	0.642060957910015\\
0.0151515151515152	0.711828737300435\\
0.0181818181818182	0.760283018867924\\
0.0212121212121212	0.795493468795356\\
0.0242424242424242	0.817931785195936\\
0.0272727272727273	0.833635703918723\\
0.0303030303030303	0.845246734397678\\
0.0333333333333333	0.852808417997097\\
0.0363636363636364	0.859789550072569\\
0.0393939393939394	0.864499274310595\\
0.0424242424242424	0.869049346879536\\
0.0454545454545455	0.872518142235123\\
0.0484848484848485	0.876153846153846\\
0.0515151515151515	0.879426705370102\\
0.0545454545454545	0.882423802612482\\
0.0575757575757576	0.884840348330914\\
0.0606060606060606	0.887119013062409\\
0.0636363636363636	0.889506531204645\\
0.0666666666666667	0.891741654571843\\
0.0696969696969697	0.893759071117562\\
0.0727272727272727	0.895253991291727\\
0.0757575757575758	0.896545718432511\\
0.0787878787878788	0.898098693759071\\
0.0818181818181818	0.899179970972424\\
0.0848484848484849	0.900391872278665\\
0.0878787878787879	0.901465892597968\\
0.0909090909090909	0.902626995645864\\
0.0939393939393939	0.903526850507983\\
0.096969696969697	0.904564586357039\\
0.1	0.905537010159652\\
0.103030303030303	0.906386066763425\\
0.106060606060606	0.907119013062409\\
0.109090909090909	0.907968069666183\\
0.112121212121212	0.908570391872279\\
0.115151515151515	0.909172714078374\\
0.118181818181818	0.909716981132075\\
0.121212121212121	0.910420899854862\\
0.124242424242424	0.910849056603774\\
0.127272727272727	0.911545718432511\\
0.13030303030303	0.912373004354136\\
0.133333333333333	0.913047895500726\\
0.136363636363636	0.913766328011611\\
0.139393939393939	0.914404934687953\\
0.142424242424242	0.915007256894049\\
0.145454545454545	0.915631349782293\\
0.148484848484848	0.916240928882438\\
0.151515151515152	0.91688679245283\\
0.154545454545455	0.917460087082729\\
0.157575757575758	0.918120464441219\\
0.160606060606061	0.918875181422351\\
0.163636363636364	0.91944847605225\\
0.166666666666667	0.920116110304789\\
0.16969696969697	0.92077648766328\\
0.172727272727273	0.921313497822932\\
0.175757575757576	0.921930333817126\\
0.178787878787879	0.922343976777939\\
0.181818181818182	0.923018867924528\\
0.184848484848485	0.923512336719884\\
0.187878787878788	0.92422351233672\\
0.190909090909091	0.924934687953556\\
0.193939393939394	0.925537010159652\\
0.196969696969697	0.926342525399129\\
0.2	0.927452830188679\\
0.203030303030303	0.928367198838897\\
};
\addlegendentry{ADD};
\addplot [color=cool_purple,solid,line width=1.5pt]
  table[row sep=crcr]{%
0	0.654462989840348\\
0.00303030303030303	0.688592162554426\\
0.00606060606060606	0.714259796806967\\
0.00909090909090909	0.735043541364296\\
0.0121212121212121	0.76699564586357\\
0.0151515151515152	0.795014513788099\\
0.0181818181818182	0.817503628447025\\
0.0212121212121212	0.83489114658926\\
0.0242424242424242	0.84766328011611\\
0.0272727272727273	0.856233671988389\\
0.0303030303030303	0.862184325108853\\
0.0333333333333333	0.866966618287373\\
0.0363636363636364	0.870573294629898\\
0.0393939393939394	0.873541364296081\\
0.0424242424242424	0.876117561683599\\
0.0454545454545455	0.878200290275762\\
0.0484848484848485	0.880377358490566\\
0.0515151515151515	0.882402031930334\\
0.0545454545454545	0.883976777939042\\
0.0575757575757576	0.88533381712627\\
0.0606060606060606	0.886937590711176\\
0.0636363636363636	0.888548621190131\\
0.0666666666666667	0.889847605224964\\
0.0696969696969697	0.891161103047896\\
0.0727272727272727	0.892743105950653\\
0.0757575757575758	0.894034833091437\\
0.0787878787878788	0.895420899854862\\
0.0818181818181818	0.896930333817127\\
0.0848484848484849	0.89800435413643\\
0.0878787878787879	0.899208998548621\\
0.0909090909090909	0.900428156748912\\
0.0939393939393939	0.901458635703919\\
0.096969696969697	0.902365747460087\\
0.1	0.903374455732947\\
0.103030303030303	0.904252539912918\\
0.106060606060606	0.905029027576197\\
0.109090909090909	0.905965166908563\\
0.112121212121212	0.906894049346879\\
0.115151515151515	0.908134978229318\\
0.118181818181818	0.908991291727141\\
0.121212121212121	0.910029027576198\\
0.124242424242424	0.910936139332366\\
0.127272727272727	0.911814223512337\\
0.13030303030303	0.912510885341074\\
0.133333333333333	0.913345428156749\\
0.136363636363636	0.913925979680696\\
0.139393939393939	0.914615384615384\\
0.142424242424242	0.915580551523948\\
0.145454545454545	0.916262699564586\\
0.148484848484848	0.916792452830189\\
0.151515151515152	0.917307692307692\\
0.154545454545455	0.918018867924528\\
0.157575757575758	0.91844702467344\\
0.160606060606061	0.918984034833091\\
0.163636363636364	0.919521044992743\\
0.166666666666667	0.919905660377358\\
0.16969696969697	0.920355587808418\\
0.172727272727273	0.920754716981132\\
0.175757575757576	0.921269956458636\\
0.178787878787879	0.921690856313498\\
0.181818181818182	0.922140783744557\\
0.184848484848485	0.922590711175617\\
0.187878787878788	0.923149492017417\\
0.190909090909091	0.923730043541364\\
0.193939393939394	0.924129172714078\\
0.196969696969697	0.9244412191582\\
0.2	0.924876632801161\\
0.203030303030303	0.925355587808418\\
};
\addlegendentry{GCNN};
\addplot [color=cool_blue,solid,line width=1.5pt]
  table[row sep=crcr]{%
0					0.624397677793904\\
0.00303030303030303	0.655348330914369\\
0.00606060606060606	0.730957910014514\\
0.00909090909090909	0.806894049346879\\
0.0121212121212121	0.876262699564586\\
0.0151515151515152	0.929303338171263\\
0.0181818181818182	0.959978229317852\\
0.0212121212121212	0.976727140783744\\
0.0242424242424242	0.98455732946299\\
0.0272727272727273	0.988795355587808\\
0.0303030303030303	0.991654571843251\\
0.0333333333333333	0.99344702467344\\
0.0363636363636364	0.994383164005805\\
0.0393939393939394	0.995145137880987\\
0.0424242424242424	0.995566037735849\\
0.0454545454545455	0.995870827285922\\
0.0484848484848485	0.996146589259797\\
0.0515151515151515	0.996320754716981\\
0.0545454545454545	0.996480406386067\\
0.0575757575757576	0.996582002902757\\
0.0606060606060606	0.996625544267054\\
0.0636363636363636	0.996683599419449\\
0.0666666666666667	0.996698113207547\\
0.0696969696969697	0.996734397677794\\
0.0727272727272727	0.996792452830189\\
0.0757575757575758	0.996806966618287\\
0.0787878787878788	0.996821480406386\\
0.0818181818181818	0.996835994194485\\
0.0848484848484849	0.996857764876633\\
0.0878787878787879	0.996865021770682\\
0.0909090909090909	0.996865021770682\\
0.0939393939393939	0.996879535558781\\
0.096969696969697		0.99688679245283\\
0.1					0.996894049346879\\
0.103030303030303	0.996894049346879\\
0.106060606060606	0.996894049346879\\
0.109090909090909	0.996894049346879\\
0.112121212121212	0.996908563134978\\
0.115151515151515	0.996908563134978\\
0.118181818181818	0.996915820029027\\
0.121212121212121	0.996930333817126\\
0.124242424242424	0.996937590711176\\
0.127272727272727	0.996944847605225\\
0.13030303030303	0.996952104499274\\
0.133333333333333	0.996959361393323\\
0.136363636363636	0.996973875181422\\
0.139393939393939	0.996973875181422\\
0.142424242424242	0.996973875181422\\
0.145454545454545	0.996988388969521\\
0.148484848484848	0.996988388969521\\
0.151515151515152	0.997010159651669\\
0.154545454545455	0.997010159651669\\
0.157575757575758	0.997010159651669\\
0.160606060606061	0.997010159651669\\
0.163636363636364	0.997017416545718\\
0.166666666666667	0.997017416545718\\
0.16969696969697	0.997024673439768\\
0.172727272727273	0.997024673439768\\
0.175757575757576	0.997024673439768\\
0.178787878787879	0.997031930333817\\
0.181818181818182	0.997031930333817\\
0.184848484848485	0.997039187227866\\
0.187878787878788	0.997039187227866\\
0.190909090909091	0.997039187227866\\
0.193939393939394	0.997039187227866\\
0.196969696969697	0.997046444121916\\
0.2				0.997046444121916\\
0.203030303030303	0.997046444121916\\
};
\addlegendentry{ACNN};

\addplot [color=cool_red,dashed,line width=3.0pt,forget plot]
  table[row sep=crcr]{%
0					0.738497822931785\\
0.00303030303030303	0.763381712626996\\
0.00606060606060606	0.79011611030479\\
0.00909090909090909	0.816640058055153\\
0.0121212121212121	0.855283018867924\\
0.0151515151515152	0.887235123367199\\
0.0181818181818182	0.912358490566038\\
0.0212121212121212	0.928628447024674\\
0.0242424242424242	0.939513788098694\\
0.0272727272727273	0.945616835994194\\
0.0303030303030303	0.949404934687954\\
0.0333333333333333	0.952140783744557\\
0.0363636363636364	0.954499274310595\\
0.0393939393939394	0.955957910014514\\
0.0424242424242424	0.957358490566038\\
0.0454545454545455	0.958243831640058\\
0.0484848484848485	0.959274310595065\\
0.0515151515151515	0.960210449927431\\
0.0545454545454545	0.960943396226415\\
0.0575757575757576	0.96155297532656\\
0.0606060606060606	0.962010159651669\\
0.0636363636363636	0.962568940493469\\
0.0666666666666667	0.962946298984035\\
0.0696969696969697	0.963258345428157\\
0.0727272727272727	0.963657474600871\\
0.0757575757575758	0.964071117561684\\
0.0787878787878788	0.96455007256894\\
0.0818181818181818	0.964992743105951\\
0.0848484848484849	0.965420899854862\\
0.0878787878787879	0.965798258345428\\
0.0909090909090909	0.966052249637155\\
0.0939393939393939	0.966407837445573\\
0.096969696969697		0.966705370101597\\
0.1					0.967010159651669\\
0.103030303030303	0.967256894049347\\
0.106060606060606	0.967597968069666\\
0.109090909090909	0.967844702467344\\
0.112121212121212	0.968156748911466\\
0.115151515151515	0.968490566037736\\
0.118181818181818	0.96877358490566\\
0.121212121212121	0.969071117561684\\
0.124242424242424	0.969208998548621\\
0.127272727272727	0.969419448476052\\
0.13030303030303	0.969680696661829\\
0.133333333333333	0.969847605224964\\
0.136363636363636	0.97000725689405\\
0.139393939393939	0.970159651669086\\
0.142424242424242	0.970297532656023\\
0.145454545454545	0.970478955007257\\
0.148484848484848	0.970609579100145\\
0.151515151515152	0.970732946298984\\
0.154545454545455	0.970899854862119\\
0.157575757575758	0.971052249637155\\
0.160606060606061	0.971204644412192\\
0.163636363636364	0.971277213352685\\
0.166666666666667	0.971436865021771\\
0.16969696969697	0.971603773584906\\
0.172727272727273	0.971698113207547\\
0.175757575757576	0.971835994194485\\
0.178787878787879	0.971952104499274\\
0.181818181818182	0.972089985486212\\
0.184848484848485	0.972235123367199\\
0.187878787878788	0.972373004354136\\
0.190909090909091	0.972568940493469\\
0.193939393939394	0.972721335268505\\
0.196969696969697	0.972822931785196\\
0.2				0.973011611030479\\
0.203030303030303	0.973214804063861\\
};
\addlegendentry{MoNet};

\addplot [color=cool_red,solid,line width=3.0pt]
  table[row sep=crcr]{%
0					0.881763425253991\\
0.00303030303030303	0.887568940493469\\
0.00606060606060606	0.897489114658926\\
0.00909090909090909	0.913396226415094\\
0.0121212121212121	0.936480406386067\\
0.0151515151515152	0.958969521044993\\
0.0181818181818182	0.975290275761974\\
0.0212121212121212	0.984158200290276\\
0.0242424242424242	0.9883381712627\\
0.0272727272727273	0.991190130624093\\
0.0303030303030303	0.993483309143687\\
0.0333333333333333	0.995609579100145\\
0.0363636363636364	0.997140783744558\\
0.0393939393939394	0.997583454281568\\
0.0424242424242424	0.998084179970973\\
0.0454545454545455	0.998272859216255\\
0.0484848484848485	0.998425253991292\\
0.0515151515151515	0.998526850507983\\
0.0545454545454545	0.998613933236575\\
0.0575757575757576	0.998751814223512\\
0.0606060606060606	0.998788098693759\\
0.0636363636363636	0.998838896952104\\
0.0666666666666667	0.998867924528302\\
0.0696969696969697	0.9988824383164\\
0.0727272727272727	0.99888969521045\\
0.0757575757575758	0.998911465892598\\
0.0787878787878788	0.998925979680697\\
0.0818181818181818	0.998947750362845\\
0.0848484848484849	0.998955007256894\\
0.0878787878787879	0.999027576197387\\
0.0909090909090909	0.999056603773585\\
0.0939393939393939	0.999092888243832\\
0.096969696969697		0.999179970972424\\
0.1					0.999238026124819\\
0.103030303030303	0.999274310595065\\
0.106060606060606	0.99933236574746\\
0.109090909090909	0.999404934687954\\
0.112121212121212	0.999455732946299\\
0.115151515151515	0.999542815674891\\
0.118181818181818	0.999680696661829\\
0.121212121212121	0.999724238026125\\
0.124242424242424	0.999760522496371\\
0.127272727272727	0.999760522496371\\
0.13030303030303	0.999767779390421\\
0.133333333333333	0.999767779390421\\
0.136363636363636	0.999767779390421\\
0.139393939393939	0.999767779390421\\
0.142424242424242	0.999767779390421\\
0.145454545454545	0.999767779390421\\
0.148484848484848	0.999767779390421\\
0.151515151515152	0.999767779390421\\
0.154545454545455	0.999767779390421\\
0.157575757575758	0.999767779390421\\
0.160606060606061	0.999767779390421\\
0.163636363636364	0.999767779390421\\
0.166666666666667	0.999767779390421\\
0.16969696969697	0.999767779390421\\
0.172727272727273	0.999767779390421\\
0.175757575757576	0.999767779390421\\
0.178787878787879	0.999767779390421\\
0.181818181818182	0.999767779390421\\
0.184848484848485	0.999767779390421\\
0.187878787878788	0.999767779390421\\
0.190909090909091	0.999767779390421\\
0.193939393939394	0.999767779390421\\
0.196969696969697	0.999767779390421\\
0.2				0.999767779390421\\
0.203030303030303	0.999767779390421\\
};
\addlegendentry{MoNet};

\end{axis}
\end{tikzpicture}

%% file: Kim_comparison_range_scans.tikz
%
%
\definecolor{cool_yellow}{rgb}{0.92900,0.69400,0.12500}%
\definecolor{cool_green}{rgb}{0.46600,0.67400,0.18800}%
\definecolor{cool_purple}{rgb}{0.49400,0.18400,0.55600}%
\definecolor{cool_blue}{rgb}{0.00000,0.44700,0.74100}%
\definecolor{cool_red}{rgb}{0.85000,0.32500,0.09800}%
\begin{tikzpicture}

\pgfplotsset{compat=newest} 

\tikzstyle{every node}=[font=\footnotesize]

\begin{axis}[%
width=0.85\figurewidth,
height=\figureheight,
scale only axis,
xmin=0,
xmax=0.2,
xlabel={Geodesic error (cm)},
xtick = {0,0.05,0.1,0.15,0.2},
xticklabels = {0,10,20,30,40},
axis x line*=top,
axis y line=none,
]
\addplot [color=white,opacity=0.0,line width=1pt,forget plot]
  table[row sep=crcr]{%
0	0.0661103047895501\\
0.203030303030303	0.998846153846154\\
};
\end{axis}

\begin{axis}[%
width=0.85\figurewidth,
height=\figureheight,
at={(0\figurewidth,0\figureheight)},
scale only axis,
xmin=0,
xmax=0.2,
xlabel style={align=center,text width=5cm},
xlabel={\vspace{-0.5cm}\begin{center}Geodesic error (\% diameter) 
\end{center}
},
xtick={0,0.05,0.1,0.15,0.2},
xticklabels = {0,0.05,0.1,0.15,0.2},
xmajorgrids,
ymin=0,
ymax=1,
ytick={0,0.2,0.4,0.6,0.8,1},
ylabel={\% correspondences},
ymajorgrids,
axis background/.style={fill=white},
legend style={
	at={(0.99,0.01)},
	anchor=south east,
	legend cell align=left,
	align=left,
	text width=6.50em,
	text height=1ex,
}
]

\addplot [color=cool_green,dashed,line width=1.5pt,forget plot]
  table[row sep=crcr]{%
0					0.185725446699647\\
0.00303030303030303	0.186023534483548\\
0.00606060606060606	0.189638493150138\\
0.00909090909090909	0.201471886260931\\
0.0121212121212121	0.23047151919219\\
0.0151515151515152	0.270107981129208\\
0.0181818181818182	0.315787022958624\\
0.0212121212121212	0.365742175287139\\
0.0242424242424242	0.415027995761537\\
0.0272727272727273	0.454385044459658\\
0.0303030303030303	0.479532793860501\\
0.0333333333333333	0.497534913988875\\
0.0363636363636364	0.510050626600633\\
0.0393939393939394	0.521286075894181\\
0.0424242424242424	0.531463340854177\\
0.0454545454545455	0.540719400333426\\
0.0484848484848485	0.548470097404602\\
0.0515151515151515	0.556202191180208\\
0.0545454545454545	0.564024146568903\\
0.0575757575757576	0.570128034239839\\
0.0606060606060606	0.575709060493571\\
0.0636363636363636	0.581120486278972\\
0.0666666666666667	0.585444189264572\\
0.0696969696969697	0.590079301326962\\
0.0727272727272727	0.594337650725814\\
0.0757575757575758	0.598434828006944\\
0.0787878787878788	0.6025925146672\\
0.0818181818181818	0.606719801050293\\
0.0848484848484849	0.610543369271526\\
0.0878787878787879	0.614114101744186\\
0.0909090909090909	0.617381144366093\\
0.0939393939393939	0.620195162626119\\
0.096969696969697	0.623254722527116\\
0.1				0.625939385626026\\
0.103030303030303	0.628527816982334\\
0.106060606060606	0.631197553693922\\
0.109090909090909	0.633665779426305\\
0.112121212121212	0.635822151484795\\
0.115151515151515	0.638110809159905\\
0.118181818181818	0.640428441556416\\
0.121212121212121	0.642449264137374\\
0.124242424242424	0.6445082621346\\
0.127272727272727	0.646617086968198\\
0.13030303030303	0.649109124249953\\
0.133333333333333	0.651449764310059\\
0.136363636363636	0.653801867878839\\
0.139393939393939	0.655902135969305\\
0.142424242424242	0.658082898104307\\
0.145454545454545	0.659996842370061\\
0.148484848484848	0.661923126889113\\
0.151515151515152	0.663907816454367\\
0.154545454545455	0.66569392841468\\
0.157575757575758	0.667542763204817\\
0.160606060606061	0.669464706574604\\
0.163636363636364	0.671301981437313\\
0.166666666666667	0.673001443920982\\
0.16969696969697	0.674674613728159\\
0.172727272727273	0.676455932866735\\
0.175757575757576	0.678464392933158\\
0.178787878787879	0.680221708599969\\
0.181818181818182	0.681814403718875\\
0.184848484848485	0.683591750854109\\
0.187878787878788	0.685306396749882\\
0.190909090909091	0.687041574745482\\
0.193939393939394	0.688659348475518\\
0.196969696969697	0.690677044985735\\
0.2	0.692505568863024\\
0.203030303030303	0.694278201079789\\
};
> CNN 3L ON DEPTH
 
> CNN 3L ON DEPTH refined
\addplot [color=cool_green,solid,line width=1.5pt]
  table[row sep=crcr]{%
0					0.156629303302702\\
0.00303030303030303	0.156841172523664\\
0.00606060606060606	0.159351067789971\\
0.00909090909090909	0.168235174264766\\
0.0121212121212121	0.194305129897017\\
0.0151515151515152	0.233953438714977\\
0.0181818181818182	0.283126997916003\\
0.0212121212121212	0.341458317203338\\
0.0242424242424242	0.403272583050636\\
0.0272727272727273	0.456558921285075\\
0.0303030303030303	0.49740640772489\\
0.0333333333333333	0.531857342392133\\
0.0363636363636364	0.562223239761336\\
0.0393939393939394	0.589857325909677\\
0.0424242424242424	0.61554104786611\\
0.0454545454545455	0.639980816987272\\
0.0484848484848485	0.660831374019259\\
0.0515151515151515	0.680814511960516\\
0.0545454545454545	0.700240832752971\\
0.0575757575757576	0.716673516175917\\
0.0606060606060606	0.731059009005233\\
0.0636363636363636	0.744815045538152\\
0.0666666666666667	0.756124448267607\\
0.0696969696969697	0.766646321282415\\
0.0727272727272727	0.77658882114044\\
0.0757575757575758	0.785320201615849\\
0.0787878787878788	0.793469807793904\\
0.0818181818181818	0.801122412990222\\
0.0848484848484849	0.808136933346782\\
0.0878787878787879	0.814650486948754\\
0.0909090909090909	0.820478134195713\\
0.0939393939393939	0.825842524083597\\
0.096969696969697		0.830811622371031\\
0.1				0.835431400960934\\
0.103030303030303	0.839705841273257\\
0.106060606060606	0.843754917587587\\
0.109090909090909	0.847455419211424\\
0.112121212121212	0.850749732019667\\
0.115151515151515	0.854080833606504\\
0.118181818181818	0.857192674975633\\
0.121212121212121	0.860214646312404\\
0.124242424242424	0.862679732333864\\
0.127272727272727	0.865147225743058\\
0.13030303030303	0.867471414119363\\
0.133333333333333	0.869780463633922\\
0.136363636363636	0.871938029527299\\
0.139393939393939	0.873901270588536\\
0.142424242424242	0.875773971497207\\
0.145454545454545	0.877213983838797\\
0.148484848484848	0.878942226879951\\
0.151515151515152	0.880528995741889\\
0.154545454545455	0.881914657131294\\
0.157575757575758	0.883409567527048\\
0.160606060606061	0.884665377282366\\
0.163636363636364	0.885847723259458\\
0.166666666666667	0.887075100293029\\
0.16969696969697	0.888116563066514\\
0.172727272727273	0.889230669993993\\
0.175757575757576	0.890233163107215\\
0.178787878787879	0.891227040627131\\
0.181818181818182	0.892086116003749\\
0.184848484848485	0.892955007296296\\
0.187878787878788	0.893689067850339\\
0.190909090909091	0.89445518880885\\
0.193939393939394	0.895188953403851\\
0.196969696969697	0.895842289468306\\
0.2				0.896576044746477\\
0.203030303030303	0.897185117995532\\
};
\addlegendentry{CNN on depth};

\addplot [color=cool_blue,dashed,line width=1.5pt,forget plot]
  table[row sep=crcr]{%
0					0.228650529722537\\
0.00303030303030303	0.228882340171733\\
0.00606060606060606	0.232261281215247\\
0.00909090909090909	0.247423636930395\\
0.0121212121212121	0.285998171762613\\
0.0151515151515152	0.338752284327589\\
0.0181818181818182	0.401194651135047\\
0.0212121212121212	0.46463397784349\\
0.0242424242424242	0.518377632797297\\
0.0272727272727273	0.562696422304376\\
0.0303030303030303	0.589983323158715\\
0.0333333333333333	0.610159519145129\\
0.0363636363636364	0.624556988474654\\
0.0393939393939394	0.636224571926929\\
0.0424242424242424	0.645126662409784\\
0.0454545454545455	0.65242498863426\\
0.0484848484848485	0.658695997482606\\
0.0515151515151515	0.664656279378075\\
0.0545454545454545	0.67116830551806\\
0.0575757575757576	0.67639932642116\\
0.0606060606060606	0.680614609897072\\
0.0636363636363636	0.68516543627888\\
0.0666666666666667	0.688707523372828\\
0.0696969696969697	0.692276327536462\\
0.0727272727272727	0.695502974039\\
0.0757575757575758	0.69890286782941\\
0.0787878787878788	0.7019961709243\\
0.0818181818181818	0.705306575144734\\
0.0848484848484849	0.708110188745514\\
0.0878787878787879	0.710319393633857\\
0.0909090909090909	0.712528127284598\\
0.0939393939393939	0.714355191180753\\
0.096969696969697		0.716377708830381\\
0.1					0.718037025915348\\
0.103030303030303	0.719677050130342\\
0.106060606060606	0.721403969579659\\
0.109090909090909	0.723020403964382\\
0.112121212121212	0.72446521902143\\
0.115151515151515	0.725841473587682\\
0.118181818181818	0.727278775641157\\
0.121212121212121	0.728639339969153\\
0.124242424242424	0.729939290662867\\
0.127272727272727	0.731436290618279\\
0.13030303030303	0.7330393805934\\
0.133333333333333	0.734709451832719\\
0.136363636363636	0.736277704553004\\
0.139393939393939	0.737961980220355\\
0.142424242424242	0.739815829928782\\
0.145454545454545	0.741515269905422\\
0.148484848484848	0.743325099550894\\
0.151515151515152	0.745064466687977\\
0.154545454545455	0.74642671507114\\
0.157575757575758	0.748195725914987\\
0.160606060606061	0.749933800507572\\
0.163636363636364	0.751550189226789\\
0.166666666666667	0.752815273296874\\
0.16969696969697	0.754141349139704\\
0.172727272727273	0.75537452710852\\
0.175757575757576	0.756914106575898\\
0.178787878787879	0.758181295285768\\
0.181818181818182	0.75937861252409\\
0.184848484848485	0.760514935881006\\
0.187878787878788	0.761575170608586\\
0.190909090909091	0.762768016199758\\
0.193939393939394	0.763828695950609\\
0.196969696969697	0.764926833145571\\
0.2				0.766014587560491\\
0.203030303030303	0.767092489749706\\
};

\addplot [color=cool_blue,solid,line width=1.5pt]
  table[row sep=crcr]{%
0					0.202764671966876\\
0.00303030303030303	0.20295641357158\\
0.00606060606060606	0.205853343226635\\
0.00909090909090909	0.218687466967991\\
0.0121212121212121	0.254161929500361\\
0.0151515151515152	0.307711715074547\\
0.0181818181818182	0.373950420122224\\
0.0212121212121212	0.445395493664996\\
0.0242424242424242	0.510845554811956\\
0.0272727272727273	0.567898572937444\\
0.0303030303030303	0.609801341346185\\
0.0333333333333333	0.645532228517175\\
0.0363636363636364	0.676522908513761\\
0.0393939393939394	0.703010514904831\\
0.0424242424242424	0.725218983755028\\
0.0454545454545455	0.74480343049978\\
0.0484848484848485	0.76252389331972\\
0.0515151515151515	0.779194219399228\\
0.0545454545454545	0.795802066409605\\
0.0575757575757576	0.809249769579802\\
0.0606060606060606	0.820231904363687\\
0.0636363636363636	0.831054251646452\\
0.0666666666666667	0.839784135995788\\
0.0696969696969697	0.847837061805634\\
0.0727272727272727	0.854998321657015\\
0.0757575757575758	0.861364677160199\\
0.0787878787878788	0.867287603505379\\
0.0818181818181818	0.872942608945379\\
0.0848484848484849	0.878118686124356\\
0.0878787878787879	0.882362104412984\\
0.0909090909090909	0.886135899956938\\
0.0939393939393939	0.889306208893042\\
0.096969696969697		0.892405218408884\\
0.1				0.895336071394115\\
0.103030303030303	0.897827264077693\\
0.106060606060606	0.900131290317517\\
0.109090909090909	0.90219838719928\\
0.112121212121212	0.904122368916717\\
0.115151515151515	0.905845579620551\\
0.118181818181818	0.907524953303712\\
0.121212121212121	0.908984895221549\\
0.124242424242424	0.910301134456252\\
0.127272727272727	0.911486791265089\\
0.13030303030303	0.912765423582197\\
0.133333333333333	0.913833633822263\\
0.136363636363636	0.914884163395035\\
0.139393939393939	0.915748464588389\\
0.142424242424242	0.916653131852404\\
0.145454545454545	0.917513063867616\\
0.148484848484848	0.918267543895559\\
0.151515151515152	0.919022364201326\\
0.154545454545455	0.919644114768338\\
0.157575757575758	0.920223035686353\\
0.160606060606061	0.920847997613097\\
0.163636363636364	0.921495315352184\\
0.166666666666667	0.922063895255541\\
0.16969696969697	0.922570471878475\\
0.172727272727273	0.923041419382175\\
0.175757575757576	0.923545020299706\\
0.178787878787879	0.924026951118277\\
0.181818181818182	0.924460377908688\\
0.184848484848485	0.924861205748305\\
0.187878787878788	0.925251556191959\\
0.190909090909091	0.925635102625503\\
0.193939393939394	0.925995906112658\\
0.196969696969697	0.92627720899\\
0.2				0.926603859494878\\
0.203030303030303	0.926887842244376\\
};
\addlegendentry{CNN on SHOT};
\addplot [color=cool_red,dashed,line width=3.0pt,forget plot]
  table[row sep=crcr]{%
0	0.408370627719179\\
0.00303030303030303	0.408657267347428\\
0.00606060606060606	0.413666943518781\\
0.00909090909090909	0.437892311232086\\
0.0121212121212121	0.494544772022521\\
0.0151515151515152	0.569252533764085\\
0.0181818181818182	0.651505019894929\\
0.0212121212121212	0.724446581588958\\
0.0242424242424242	0.782101706440681\\
0.0272727272727273	0.825617791236501\\
0.0303030303030303	0.849579694913194\\
0.0333333333333333	0.864680613117821\\
0.0363636363636364	0.873869613534966\\
0.0393939393939394	0.881404820843455\\
0.0424242424242424	0.886783309214408\\
0.0454545454545455	0.890541030215261\\
0.0484848484848485	0.893615324399433\\
0.0515151515151515	0.896152761566898\\
0.0545454545454545	0.898719126416659\\
0.0575757575757576	0.90068746217355\\
0.0606060606060606	0.902493957069216\\
0.0636363636363636	0.904179858574256\\
0.0666666666666667	0.905430528864379\\
0.0696969696969697	0.90662149472682\\
0.0727272727272727	0.907609241371448\\
0.0757575757575758	0.908657753939788\\
0.0787878787878788	0.909603852401356\\
0.0818181818181818	0.910520471463346\\
0.0848484848484849	0.911301820897791\\
0.0878787878787879	0.911937216650909\\
0.0909090909090909	0.912580374986417\\
0.0939393939393939	0.913176469978935\\
0.096969696969697	0.913754661832543\\
0.1	0.914270041143274\\
0.103030303030303	0.914746598188425\\
0.106060606060606	0.915158416201031\\
0.109090909090909	0.915611453340045\\
0.112121212121212	0.916004254450333\\
0.115151515151515	0.916475133448865\\
0.118181818181818	0.916915505542587\\
0.121212121212121	0.917307654927924\\
0.124242424242424	0.917701918493575\\
0.127272727272727	0.918159332417137\\
0.13030303030303	0.918676529606055\\
0.133333333333333	0.91917139563778\\
0.136363636363636	0.919689955425372\\
0.139393939393939	0.920168706446499\\
0.142424242424242	0.920819423151516\\
0.145454545454545	0.921409266302896\\
0.148484848484848	0.921972045098421\\
0.151515151515152	0.922512002075549\\
0.154545454545455	0.92305380063747\\
0.157575757575758	0.923607264288809\\
0.160606060606061	0.924139926353199\\
0.163636363636364	0.924677409333945\\
0.166666666666667	0.925101256860754\\
0.16969696969697	0.925561440031671\\
0.172727272727273	0.926042257960151\\
0.175757575757576	0.92645827578292\\
0.178787878787879	0.926852024338775\\
0.181818181818182	0.927127098248151\\
0.184848484848485	0.927378116482632\\
0.187878787878788	0.92764578362053\\
0.190909090909091	0.927911582893511\\
0.193939393939394	0.928170579628771\\
0.196969696969697	0.928410450749994\\
0.2	0.928612485838958\\
0.203030303030303	0.928873783478433\\
};

\addplot [color=cool_red,solid,line width=3.0pt]
  table[row sep=crcr]{%
0	0.390889533149985\\
0.00303030303030303	0.391185207136256\\
0.00606060606060606	0.39549392522673\\
0.00909090909090909	0.417761386844243\\
0.0121212121212121	0.472157179447104\\
0.0151515151515152	0.54636312209765\\
0.0181818181818182	0.629351581020332\\
0.0212121212121212	0.704894498654206\\
0.0242424242424242	0.766604272336185\\
0.0272727272727273	0.815349916297488\\
0.0303030303030303	0.845792951149432\\
0.0333333333333333	0.867880110525343\\
0.0363636363636364	0.884188904866235\\
0.0393939393939394	0.898580465069125\\
0.0424242424242424	0.909793928852119\\
0.0454545454545455	0.918830704441524\\
0.0484848484848485	0.926526918601316\\
0.0515151515151515	0.933139842617017\\
0.0545454545454545	0.939530596580261\\
0.0575757575757576	0.94464675892047\\
0.0606060606060606	0.948932913595948\\
0.0636363636363636	0.952916311419345\\
0.0666666666666667	0.955914967804957\\
0.0696969696969697	0.958610583249153\\
0.0727272727272727	0.960825207899753\\
0.0757575757575758	0.962736895597603\\
0.0787878787878788	0.964569551441538\\
0.0818181818181818	0.966499880240762\\
0.0848484848484849	0.967982387995336\\
0.0878787878787879	0.969229807375908\\
0.0909090909090909	0.970361475966655\\
0.0939393939393939	0.971288024940208\\
0.096969696969697	0.972256799673878\\
0.1	0.973140688663762\\
0.103030303030303	0.973951021265374\\
0.106060606060606	0.974638778097915\\
0.109090909090909	0.975346935647354\\
0.112121212121212	0.975960254172196\\
0.115151515151515	0.976514070900013\\
0.118181818181818	0.977067131293172\\
0.121212121212121	0.97754155017452\\
0.124242424242424	0.977922790275951\\
0.127272727272727	0.978315320092399\\
0.13030303030303	0.978631924201\\
0.133333333333333	0.979002589712218\\
0.136363636363636	0.979316540300662\\
0.139393939393939	0.979580095221407\\
0.142424242424242	0.979906064920623\\
0.145454545454545	0.980136107848673\\
0.148484848484848	0.980338183740563\\
0.151515151515152	0.980518067238846\\
0.154545454545455	0.980747070068846\\
0.157575757575758	0.980919531962497\\
0.160606060606061	0.981058678253135\\
0.163636363636364	0.981229169068382\\
0.166666666666667	0.981382258075152\\
0.16969696969697	0.981504974471381\\
0.172727272727273	0.981642623018533\\
0.175757575757576	0.981765054618107\\
0.178787878787879	0.981853167692299\\
0.181818181818182	0.981941216367571\\
0.184848484848485	0.982022788762691\\
0.187878787878788	0.98211774965124\\
0.190909090909091	0.98220249009883\\
0.193939393939394	0.982277425672161\\
0.196969696969697	0.982361904415547\\
0.2	0.982421506070372\\
0.203030303030303	0.982497197276045\\
};
\addlegendentry{MoNet};

\end{axis}
\end{tikzpicture}

%% file: geod_errs_all.tex

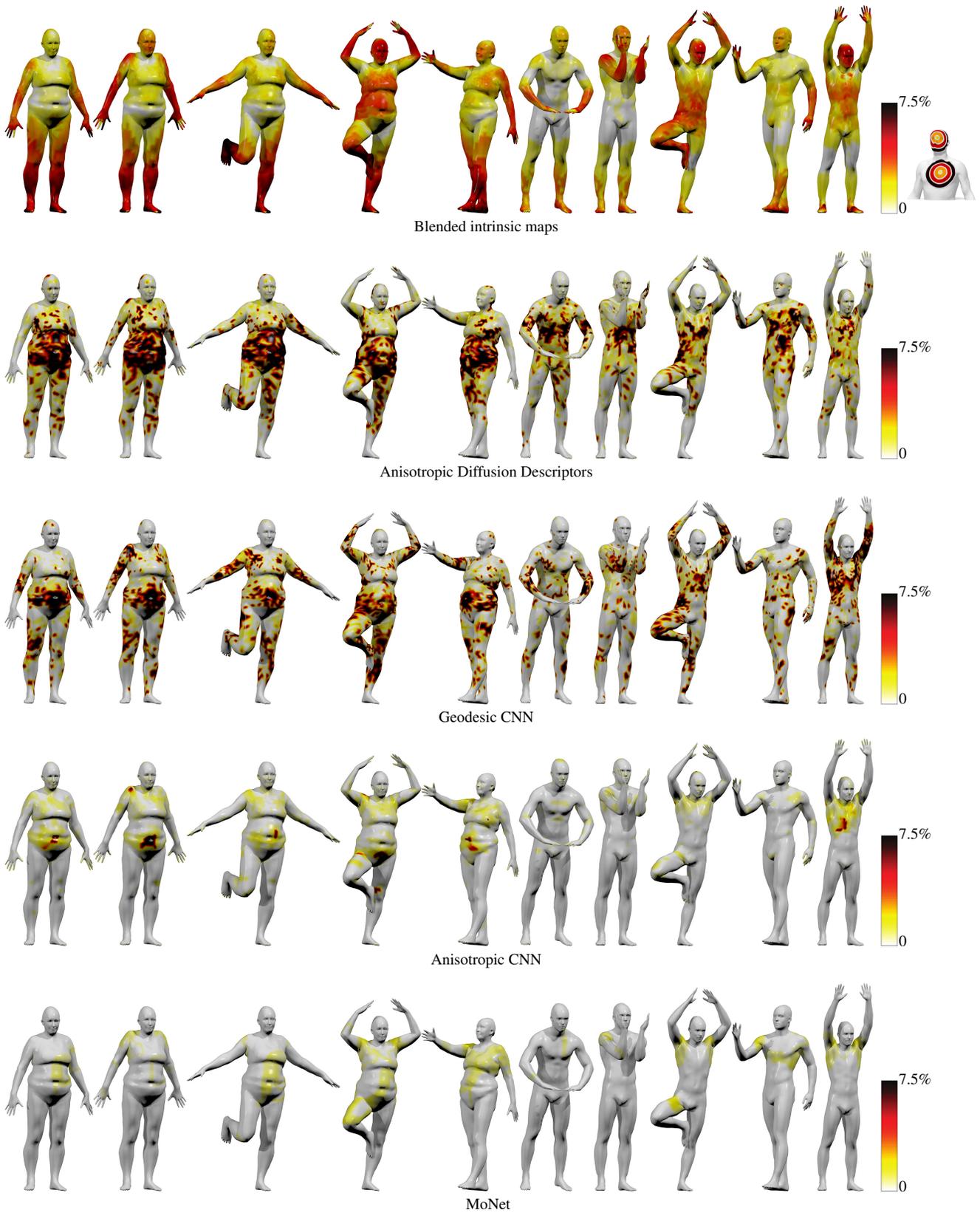
\begin{figure*}[t!]
\begin{minipage}{0.92\textwidth}
	\begin{overpic}
	[width=1\linewidth]{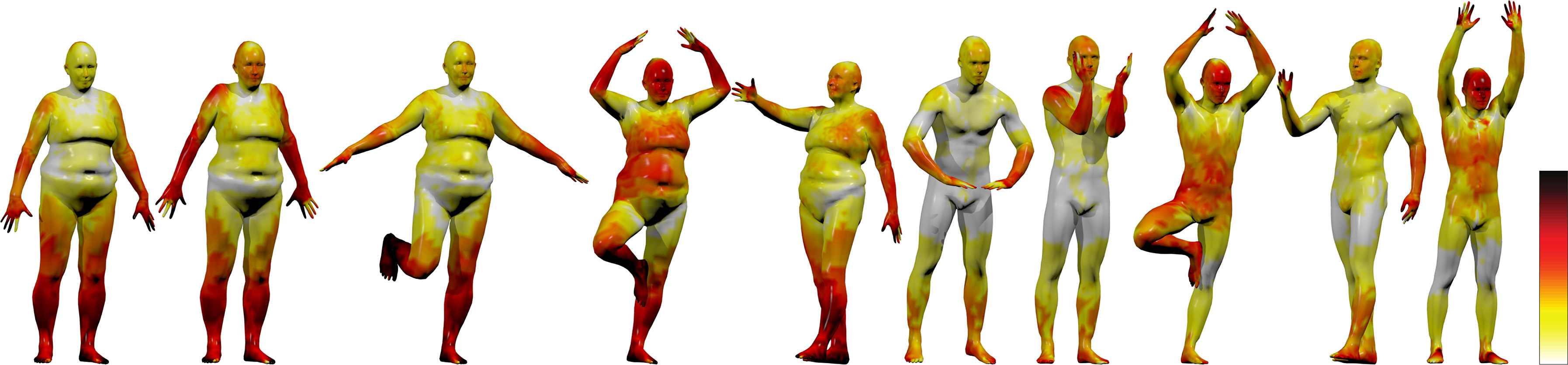}
	\put(100.15,0){\footnotesize{0}}
	\put(100.15,12){\footnotesize{7.5\%}} 
	\end{overpic}	
\end{minipage}	\hspace{0.25mm}
\begin{minipage}[t][][b]{0.07\textwidth}
\vspace{2.5mm}
		\begin{overpic}
	[width=1\linewidth]{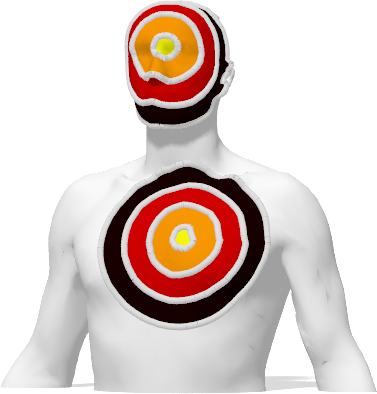}
	\end{overpic}
\end{minipage}\\
\begin{minipage}{1\textwidth}
\centering
	\vspace{1mm}
\footnotesize{Blended intrinsic maps }
	\vspace{3mm}
\end{minipage}
\begin{minipage}{0.92\textwidth}
	\begin{overpic}
	[width=1\linewidth]{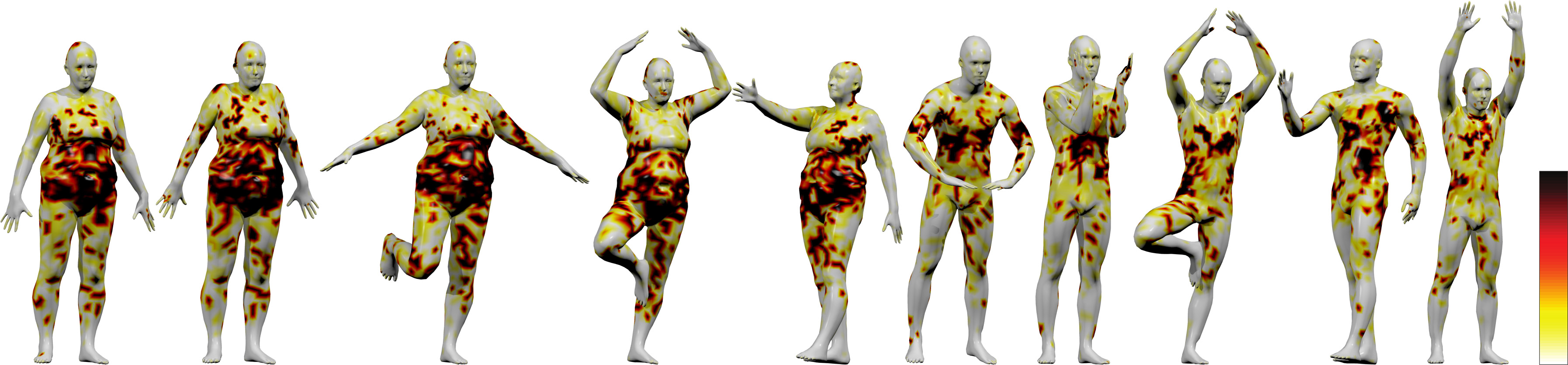}
	\put(100.15,0){\footnotesize{0}}
	\put(100.15,12){\footnotesize{7.5\%}} 
	\end{overpic}		
\end{minipage}\\
\begin{minipage}{1\textwidth}
	\centering
		\vspace{1mm}
	\footnotesize{Anisotropic Diffusion Descriptors }
		\vspace{3mm}
\end{minipage}
\begin{minipage}{0.92\textwidth}
	\begin{overpic}
	[width=1\linewidth]{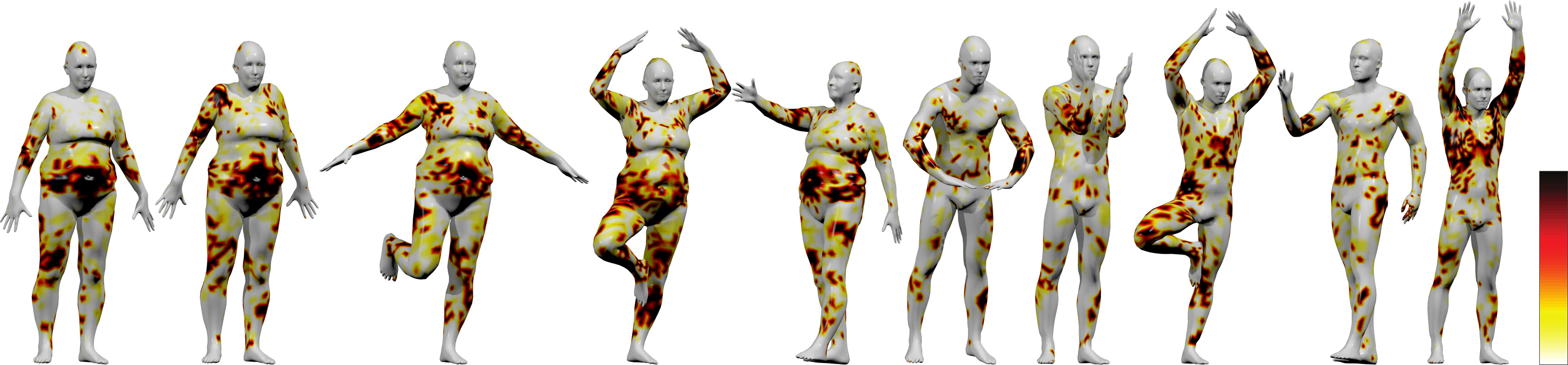}
	\put(100.15,0){\footnotesize{0}}
	\put(100.15,12){\footnotesize{7.5\%}} 
	\end{overpic}		
\end{minipage}\\
\begin{minipage}{1\textwidth}
	\centering
	\vspace{1mm}
	\footnotesize{Geodesic CNN }
		\vspace{3mm}
\end{minipage}
\begin{minipage}{0.92\textwidth}
	\begin{overpic}
	[width=1\linewidth]{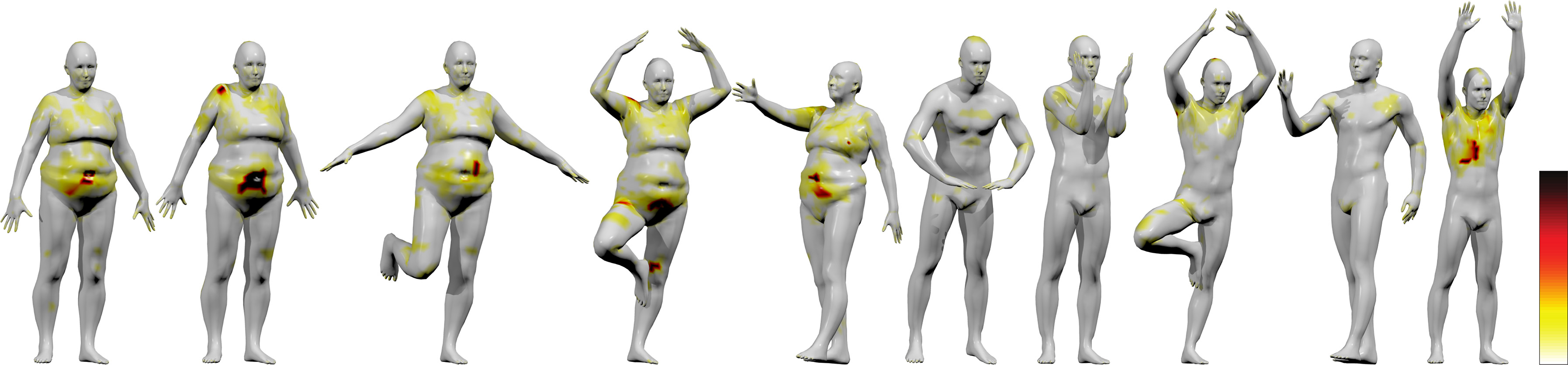}
	\put(100.15,0){\footnotesize{0}}
	\put(100.15,12){\footnotesize{7.5\%}} 
	\end{overpic}		
\end{minipage}\\
\begin{minipage}{1\textwidth}
	\centering
		\vspace{1mm}
	\footnotesize{Anisotropic CNN }
		\vspace{3mm}
\end{minipage}
\begin{minipage}{0.92\textwidth}
	\begin{overpic}
	[width=1\linewidth]{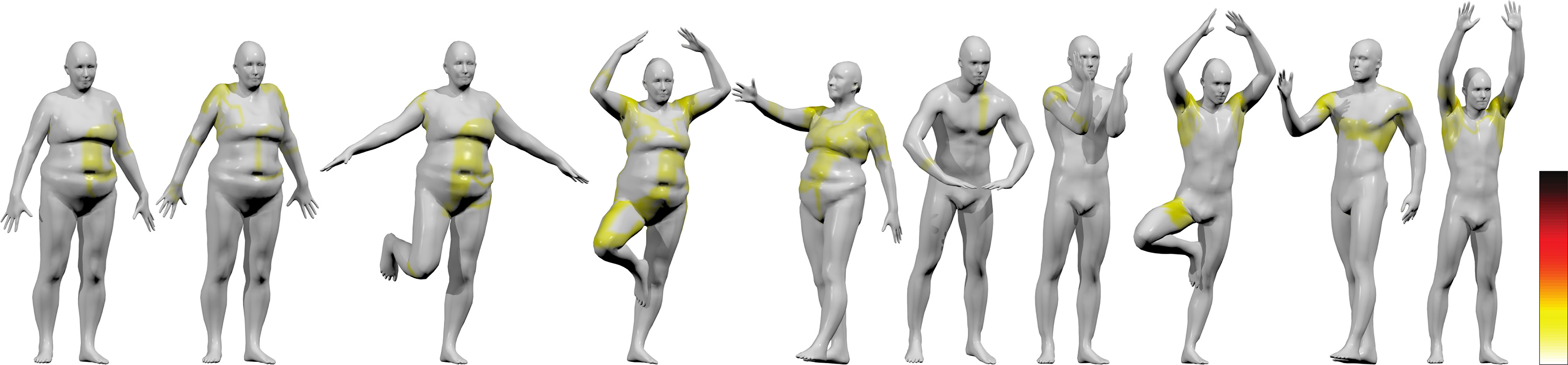}
	\put(100.15,0){\footnotesize{0}}
	\put(100.15,12){\footnotesize{7.5\%}} 
	\end{overpic}		
\end{minipage}\\
\begin{minipage}{1\textwidth}
	\centering
		\vspace{1mm}
	\footnotesize{MoNet}
		\vspace{2mm}
\end{minipage}
\vspace{0.1mm}
\caption{Pointwise error (geodesic distance from groundtruth) of different correspondence methods on the FAUST humans dataset. For visualization clarity, the error values are saturated at $7.5\%$ of the geodesic diameter, which corresponds to approximately $15$ cm. Hot colors represent large errors.}
\label{fig:geod_errs}
\end{figure*}

%% file: corrs_as_rgb.tex
\begin{figure*}[t!]
\begin{minipage}{0.95\textwidth}
	\centering 
	\begin{overpic}
	[width=1.0\textwidth]{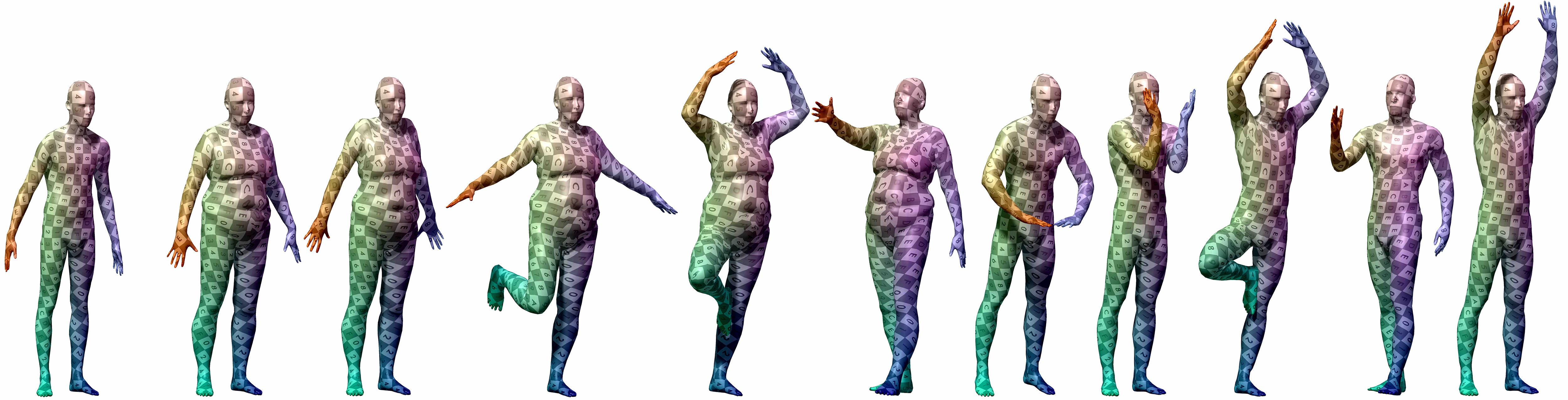}
	\end{overpic}
\end{minipage}
\vspace{0.1mm}
\caption{Examples of correspondence on the FAUST humans dataset obtained by the proposed MoNet method. Shown is the texture transferred from the leftmost reference shape to different subjects in different poses by means of our correspondence. 
}\vspace{-4mm}
\label{fig:corrs_as_rgb}
\end{figure*}

%% file: geod_errs_range_scans_all.tex

\begin{figure*}[t!]
\begin{minipage}{1.0\textwidth}
	\centering
	\begin{overpic}
	[width=1\linewidth]{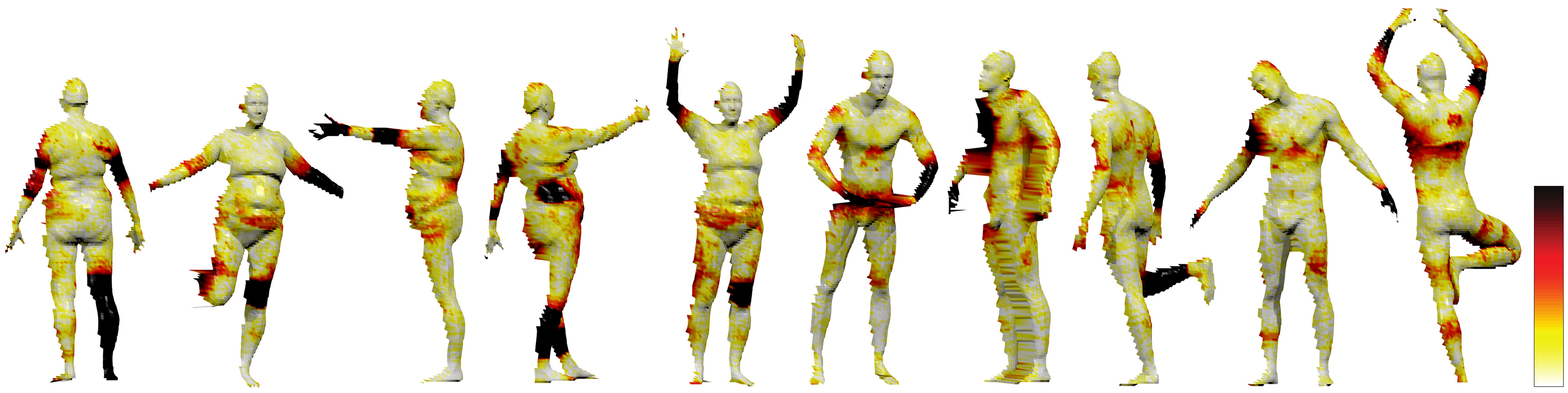}
	\put(100,0){\footnotesize{0}}
	\put(100,12.6){\footnotesize{7.5\%}}
	\end{overpic}	
\end{minipage}
\begin{minipage}{1.0\textwidth}
	\centering
	\footnotesize{Euclidean CNN}
\end{minipage}
\begin{minipage}{1.0\textwidth}
\begin{overpic}
	[width=1\linewidth]{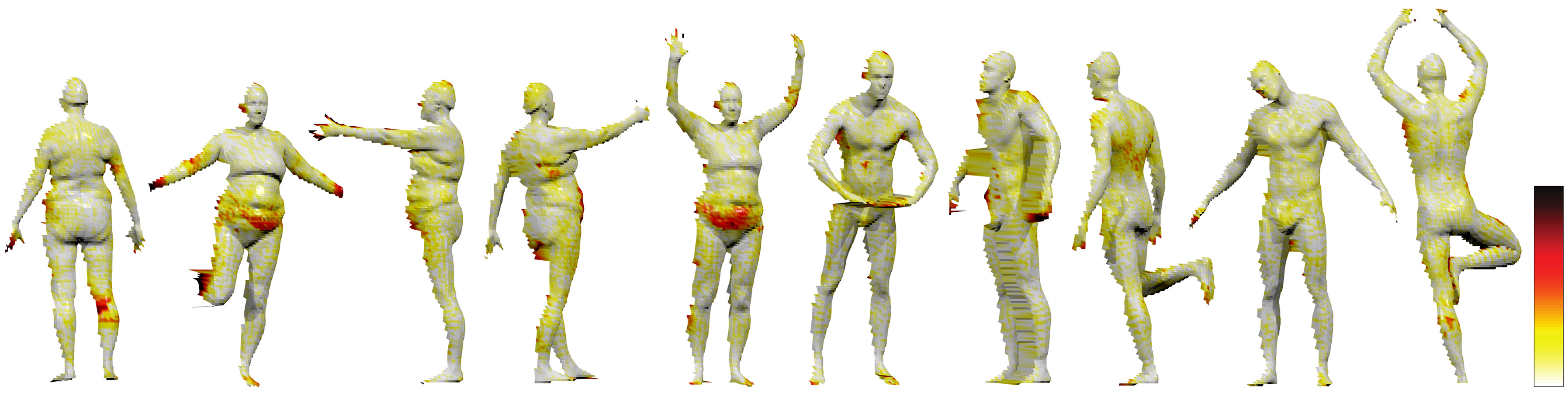}
	\put(100,0){\footnotesize{0}}
	\put(100,12.6){\footnotesize{7.5\%}}
	\end{overpic}		
\end{minipage}\\
\begin{minipage}{1.0\textwidth}
	\centering
	\footnotesize{MoNet}
\end{minipage}
\vspace{0.1mm}
\caption{Pointwise error (geodesic distance from groundtruth) of different methods on FAUST range maps. For visualization clarity, the error values are saturated at $7.5\%$ of the geodesic diameter, which corresponds to approximately $15$ cm. Hot colors represent large errors.}
\label{fig:geod_errs_range_maps}
\end{figure*}

%% file: corrs_as_rgb_range_scans.tex
\begin{figure*}[t!]
\begin{minipage}{1\textwidth}
	\centering 
	\begin{overpic}
	[width=1.0\textwidth]{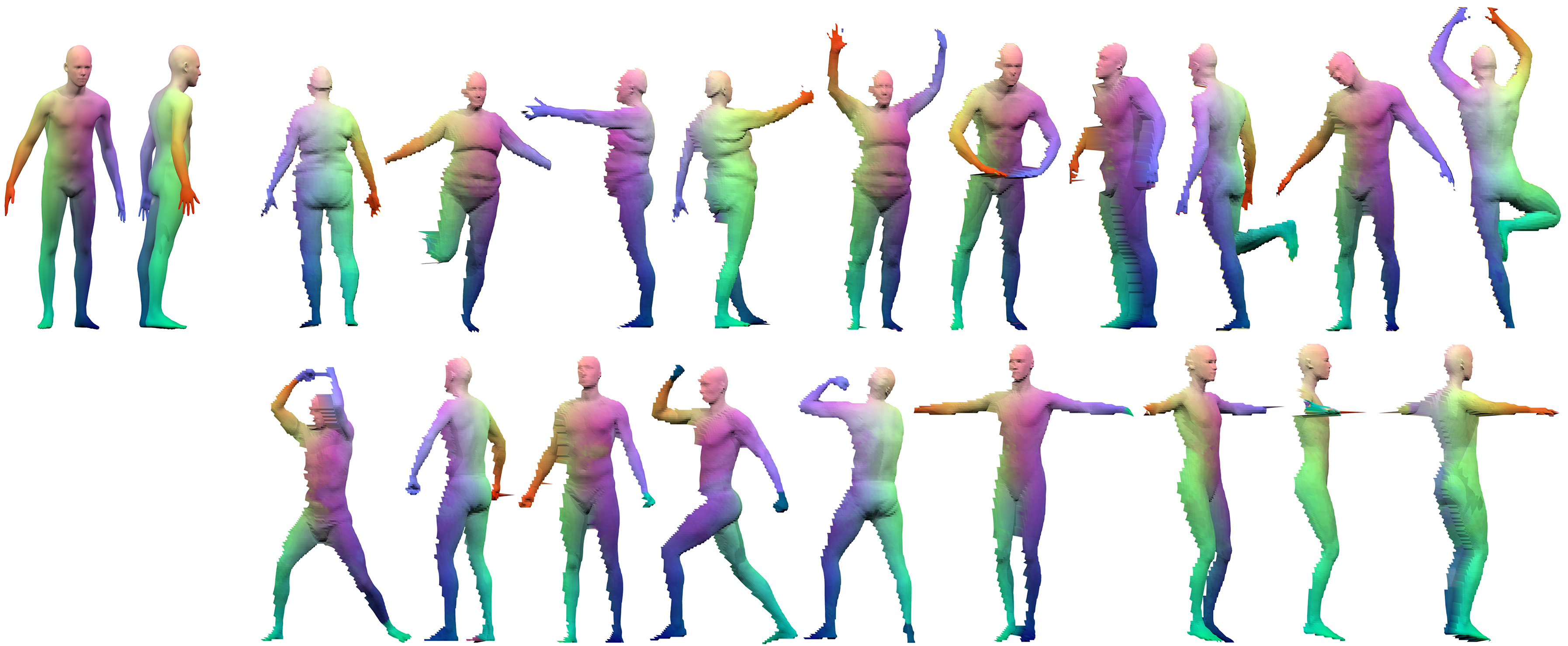}
	\end{overpic}
\end{minipage}
\vspace{2mm}
\caption{Visualization of correspondence on FAUST range maps as color code (corresponding points are shown in the same color). Full reference shape is shown on the left. Bottom row show examples of additional shapes from SCAPE and TOSCA datasets. 
}
\label{fig:corrs_as_rgb_range_scans}
\vspace{-4mm}
\end{figure*}

%% file: conclusions.tex
\section{Conclusions}

We proposed a spatial-domain model for deep learning on non-Euclidean domains such as manifolds and graphs. Our approach generalizes several previous techniques that can be obtained as particular instances thereof. Extensive experimental results show that our model is applicable to different geometric deep learning tasks, achieving state-of-the-art results. 
In deformable 3D shape analysis applications, the key advantage of our approach is that it is intrinsic and thus deformation-invariant {\em by construction}, as opposed to  Euclidean models \cite{su2015multi,wu20153d,wei2016dense,qi2016volumetric} that in general require significantly higher complexity and huge training sets to {\em learn} the deformation invariance. 
%
In future works, we will study additional promising applications of our model, for example in the domain of computational social sciences. 


%% file: main.bbl
\begin{thebibliography}{10}\itemsep=-1pt

\bibitem{anguelov2005scape}
D.~Anguelov, P.~Srinivasan, D.~Koller, S.~Thrun, J.~Rodgers, and J.~Davis.
\newblock {SCAPE}: shape completion and animation of people.
\newblock In {\em TOG}, volume~24, pages 408--416, 2005.

\bibitem{atwood2016search}
J.~Atwood and D.~Towsley.
\newblock Diffusion-convolutional neural networks.
\newblock {\em arXiv:1511.02136v2}, 2016.

\bibitem{belkin2003laplacian}
M.~Belkin and P.~Niyogi.
\newblock Laplacian eigenmaps for dimensionality reduction and data
  representation.
\newblock {\em Neural Computation}, 15(6):1373--1396, 2003.

\bibitem{belkin2006manifold}
M.~Belkin, P.~Niyogi, and V.~Sindhwani.
\newblock Manifold regularization: A geometric framework for learning from
  labeled and unlabeled examples.
\newblock {\em JMLR}, 7:2399--2434, 2006.

\bibitem{bogo}
F.~Bogo, J.~Romero, M.~Loper, and M.~J. Black.
\newblock {FAUST}: Dataset and evaluation for {3D} mesh registration.
\newblock In {\em Proc. CVPR}, 2014.

\bibitem{boscaini2015learning}
D.~Boscaini, J.~Masci, S.~Melzi, M.~M. Bronstein, U.~Castellani, and
  P.~Vandergheynst.
\newblock Learning class-specific descriptors for deformable shapes using
  localized spectral convolutional networks.
\newblock {\em Computer Graphics Forum}, 34(5):13--23, 2015.

\bibitem{boscaini2016learning}
D.~Boscaini, J.~Masci, E.~Rodol{\`a}, and M.~M. Bronstein.
\newblock Learning shape correspondence with anisotropic convolutional neural
  networks.
\newblock In {\em Proc. NIPS}, 2016.

\bibitem{boscaini2016anisotropic}
D.~Boscaini, J.~Masci, E.~Rodol{\`a}, M.~M. Bronstein, and D.~Cremers.
\newblock Anisotropic diffusion descriptors.
\newblock {\em Computer Graphics Forum}, 35(2):431--441, 2016.

\bibitem{bresson2015enhanced}
X.~Bresson, T.~Laurent, and J.~von Brecht.
\newblock Enhanced lasso recovery on graph.
\newblock In {\em Proc. EUSIPCO}, 2015.

\bibitem{bronstein2008numerical}
A.~M. Bronstein, M.~M. Bronstein, and R.~Kimmel.
\newblock {\em Numerical geometry of non-rigid shapes}.
\newblock Springer, 2008.

\bibitem{review_new}
M.~M. Bronstein, J.~Bruna, Y.~LeCun, A.~Szlam, and P.~Vandergheynst.
\newblock Geometric deep learning: going beyond euclidean data.
\newblock 2016.
\newblock preprint.

\bibitem{bruna2013spectral}
J.~Bruna, W.~Zaremba, A.~Szlam, and Y.~LeCun.
\newblock Spectral networks and locally connected networks on graphs.
\newblock {\em Proc. ICLR}, 2013.

\bibitem{cao2015grarep}
S.~Cao, W.~Lu, and Q.~Xu.
\newblock {GraRep}: Learning graph representations with global structural
  information.
\newblock In {\em Proc. IKM}, 2015.

\bibitem{coifman2006diffusionw}
R.~R. Coifman and M.~Maggioni.
\newblock Diffusion wavelets.
\newblock {\em Applied and Computational Harmonic Analysis}, 21(1):53--94,
  2006.

\bibitem{defferrard2016convolutional}
M.~Defferrard, X.~Bresson, and P.~Vandergheynst.
\newblock Convolutional neural networks on graphs with fast localized spectral
  filtering.
\newblock In {\em Proc. NIPS}, 2016.

\bibitem{dhillon2007weighted}
I.~S. Dhillon, Y.~Guan, and B.~Kulis.
\newblock Weighted graph cuts without eigenvectors: a multilevel approach.
\newblock {\em PAMI}, 29(11):1944--1957, 2007.

\bibitem{duvenaud2015convolutional}
D.~K. Duvenaud, D.~Maclaurin, J.~Iparraguirre, R.~Bombarell, T.~Hirzel,
  A.~Aspuru-Guzik, and R.~P. Adams.
\newblock Convolutional networks on graphs for learning molecular fingerprints.
\newblock In {\em Proc. NIPS}, 2015.

\bibitem{gavish2010multiscale}
M.~Gavish, B.~Nadler, and R.~R. Coifman.
\newblock Multiscale wavelets on trees, graphs and high dimensional data:
  Theory and applications to semi supervised learning.
\newblock In {\em Proc. ICML}, 2010.

\bibitem{gori2005new}
M.~Gori, G.~Monfardini, and F.~Scarselli.
\newblock A new model for learning in graph domains.
\newblock In {\em Proc. IJCNN}, 2005.

\bibitem{grovernode2vec}
A.~Grover and J.~Leskovec.
\newblock node2vec: Scalable feature learning for networks.
\newblock In {\em Proc. KDD}, 2016.

\bibitem{hammond2011wavelets}
D.~K. Hammond, P.~Vandergheynst, and R.~Gribonval.
\newblock Wavelets on graphs via spectral graph theory.
\newblock {\em Applied and Comp. Harmonic Analysis}, 30(2):129--150, 2011.

\bibitem{henaff2015deep}
M.~Henaff, J.~Bruna, and Y.~LeCun.
\newblock Deep convolutional networks on graph-structured data.
\newblock {\em arXiv:1506.05163}, 2015.

\bibitem{kalofolias2014matrix}
V.~Kalofolias, X.~Bresson, M.~M. Bronstein, and P.~Vandergheynst.
\newblock Matrix completion on graphs.
\newblock {\em arXiv:1408.1717}, 2014.

\bibitem{kim2011blended}
V.~Kim, Y.~Lipman, and T.~Funkhouser.
\newblock Blended intrinsic maps.
\newblock {\em ACM Trans. Graphics}, 30(4):79, 2011.

\bibitem{KingmaB14}
D.~P. Kingma and J.~Ba.
\newblock Adam: {A} method for stochastic optimization.
\newblock {\em arXiv:1412.6980}, 2014.

\bibitem{welling2016}
T.~N. Kipf and M.~Welling.
\newblock Semi-supervised classification with graph convolutional networks.
\newblock {\em arXiv:1609.02907}, 2016.

\bibitem{isc}
I.~Kokkinos, M.~Bronstein, R.~Litman, and A.~Bronstein.
\newblock Intrinsic shape context descriptors for deformable shapes.
\newblock In {\em Proc. CVPR}, 2012.

\bibitem{lecun1998gradient}
Y.~LeCun, L.~Bottou, Y.~Bengio, and P.~Haffner.
\newblock Gradient-based learning applied to document recognition.
\newblock {\em Proc. IEEE}, 86(11):2278--2324, 1998.

\bibitem{lezoray2012image}
O.~L{\'e}zoray and L.~Grady.
\newblock {\em Image processing and analysis with graphs: theory and practice}.
\newblock CRC Press, 2012.

\bibitem{GGSNN}
Y.~Li, D.~Tarlow, M.~Brockschmidt, and R.~Zemel.
\newblock Gated graph sequence neural networks.
\newblock {\em arXiv:1511.05493}, 2015.

\bibitem{LinCY13}
M.~Lin, Q.~Chen, and S.~Yan.
\newblock Network in network.
\newblock {\em CoRR}, abs/1312.4400, 2013.

\bibitem{masci2015geodesic}
J.~Masci, D.~Boscaini, M.~M. Bronstein, and P.~Vandergheynst.
\newblock Geodesic convolutional neural networks on {R}iemannian manifolds.
\newblock In {\em Proc. 3DRR}, 2015.

\bibitem{mikolov2013distributed}
T.~Mikolov and J.~Dean.
\newblock Distributed representations of words and phrases and their
  compositionality.
\newblock {\em Proc. NIPS}, 2013.

\bibitem{ng2002spectral}
A.~Y. Ng, M.~I. Jordan, and Y.~Weiss.
\newblock On spectral clustering: Analysis and an algorithm.
\newblock In {\em Proc. NIPS}, 2002.

\bibitem{perona1990scale}
P.~Perona and J.~Malik.
\newblock Scale-space and edge detection using anisotropic diffusion.
\newblock {\em Trans. PAMI}, 12(7):629--639, 1990.

\bibitem{perozzi2014deepwalk}
B.~Perozzi, R.~Al-Rfou, and S.~Skiena.
\newblock {DeepWalk}: Online learning of social representations.
\newblock In {\em Proc. KDD}, 2014.

\bibitem{qi2016volumetric}
C.~R. Qi, H.~Su, M.~Nie{\ss}ner, A.~Dai, M.~Yan, and L.~J. Guibas.
\newblock Volumetric and multi-view {CNN}s for object classification on {3D}
  data.
\newblock In {\em Proc. CVPR}, 2016.

\bibitem{rodola2014dense}
E.~Rodol{\`a}, S.~Rota~Bul{\`o}, T.~Windheuser, M.~Vestner, and D.~Cremers.
\newblock Dense non-rigid shape correspondence using random forests.
\newblock In {\em Proc. CVPR}, 2014.

\bibitem{rustamov2013wavelets}
R.~Rustamov and L.~J. Guibas.
\newblock Wavelets on graphs via deep learning.
\newblock In {\em Advances in Neural Information Processing Systems}, pages
  998--1006, 2013.

\bibitem{sanfeliu2002graph}
A.~Sanfeliu et~al.
\newblock Graph-based representations and techniques for image processing and
  image analysis.
\newblock {\em Pattern Recognition}, 35(3):639--650, 2002.

\bibitem{GNN}
F.~Scarselli, M.~Gori, A.~C. Tsoi, M.~Hagenbuchner, and G.~Monfardini.
\newblock The graph neural network model.
\newblock {\em {IEEE} Trans. Neural Networks}, 20(1):61--80, 2009.

\bibitem{aimag08}
P.~Sen, G.~M. Namata, M.~Bilgic, L.~Getoor, B.~Gallagher, and T.~Eliassi-Rad.
\newblock Collective classification in network data.
\newblock {\em AI Magazine}, 29(3):93--106, 2008.

\bibitem{shahid2015robust}
N.~Shahid, V.~Kalofolias, X.~Bresson, M.~M. Bronstein, and P.~Vandergheynst.
\newblock Robust principal component analysis on graphs.
\newblock In {\em Proc. ICCV}, 2015.

\bibitem{shahid2016fast}
N.~Shahid, N.~Perraudin, V.~Kalofolias, G.~Puy, and P.~Vandergheynst.
\newblock Fast robust {PCA} on graphs.
\newblock {\em IEEE J. Selected Topics in Signal Processing}, 10(4):740--756,
  2016.

\bibitem{sharon2015class}
N.~Sharon and Y.~Shkolnisky.
\newblock A class of {L}aplacian multiwavelets bases for high-dimensional data.
\newblock {\em Applied and Comp. Harmonic Analysis}, 38(3):420--451, 2015.

\bibitem{shi2000normalized}
J.~Shi and J.~Malik.
\newblock Normalized cuts and image segmentation.
\newblock {\em Trans. PAMI}, 22(8):888--905, 2000.

\bibitem{shuman2013emerging}
D.~I. Shuman, S.~K. Narang, P.~Frossard, A.~Ortega, and P.~Vandergheynst.
\newblock The emerging field of signal processing on graphs: Extending
  high-dimensional data analysis to networks and other irregular domains.
\newblock {\em {IEEE} Sig. Proc. Magazine}, 30(3):83--98, 2013.

\bibitem{shuman2016vertex}
D.~I. Shuman, B.~Ricaud, and P.~Vandergheynst.
\newblock Vertex-frequency analysis on graphs.
\newblock {\em App. and Comp. Harmonic Analysis}, 40(2):260--291, 2016.

\bibitem{sinha2016deep}
A.~Sinha, J.~Bai, and K.~Ramani.
\newblock Deep learning {3D} shape surfaces using geometry images.
\newblock In {\em Proc. ECCV}, 2016.

\bibitem{sochen1998general}
N.~Sochen, R.~Kimmel, and R.~Malladi.
\newblock A general framework for low level vision.
\newblock {\em Trans. Image Processing}, 7(3):310--318, 1998.

\bibitem{su2015multi}
H.~Su, S.~Maji, E.~Kalogerakis, and E.~Learned-Miller.
\newblock Multi-view convolutional neural networks for {3D} shape recognition.
\newblock In {\em Proc. ICCV}, 2015.

\bibitem{comnets}
S.~Sukhbaatar, A.~Szlam, and R.~Fergus.
\newblock Learning multiagent communication with backpropagation.
\newblock {\em arXiv:1605.07736}, 2016.

\bibitem{tang2015line}
J.~Tang, M.~Qu, M.~Wang, M.~Zhang, J.~Yan, and Q.~Mei.
\newblock {LINE}: Large-scale information network embedding.
\newblock In {\em Proc. WWW}, 2015.

\bibitem{tombari2010unique}
F.~Tombari, S.~Salti, and L.~Di~Stefano.
\newblock Unique signatures of histograms for local surface description.
\newblock In {\em Proc. ECCV}, 2010.

\bibitem{bayesian}
M.~Vestner, R.~Litman, A.~Bronstein, E.~Rodol\`{a}, and D.~Cremers.
\newblock Bayesian inference of bijective non-rigid shape correspondence.
\newblock {\em arXiv:1607.03425}, 2016.

\bibitem{wei2016dense}
L.~Wei, Q.~Huang, D.~Ceylan, E.~Vouga, and H.~Li.
\newblock Dense human body correspondences using convolutional networks.
\newblock In {\em Proc. CVPR}, 2016.

\bibitem{weiss2009spectral}
Y.~Weiss, A.~Torralba, and R.~Fergus.
\newblock Spectral hashing.
\newblock In {\em Proc. NIPS}, 2009.

\bibitem{weston2012deep}
J.~Weston, F.~Ratle, H.~Mobahi, and R.~Collobert.
\newblock Deep learning via semi-supervised embedding.
\newblock In {\em Neural Networks: Tricks of the Trade}, pages 639--655. 2012.

\bibitem{wu20153d}
Z.~Wu, S.~Song, A.~Khosla, F.~Yu, L.~Zhang, X.~Tang, and J.~Xiao.
\newblock {3D} shapenets: A deep representation for volumetric shapes.
\newblock In {\em Proc. CVPR}, 2015.

\bibitem{yang2016revisiting}
Z.~Yang, W.~Cohen, and R.~Salakhutdinov.
\newblock Revisiting semi-supervised learning with graph embeddings.
\newblock {\em arXiv:1603.08861}, 2016.

\bibitem{zhang2008graph}
F.~Zhang and E.~R. Hancock.
\newblock Graph spectral image smoothing using the heat kernel.
\newblock {\em Pattern Recognition}, 41(11):3328--3342, 2008.

\bibitem{zhang2012learning}
X.~Zhang, X.~Dong, and P.~Frossard.
\newblock Learning of structured graph dictionaries.
\newblock In {\em Proc. ICASSP}, 2012.

\bibitem{zhu2003semi}
X.~Zhu, Z.~Ghahramani, J.~Lafferty, et~al.
\newblock Semi-supervised learning using gaussian fields and harmonic
  functions.
\newblock In {\em Proc. ICML}, 2003.

\end{thebibliography}
